# Graduality in Argumentation

**Claudette Cayrol**                                                         CCAYROL@IRIT.FR
**Marie-Christine Lagasquie-Schiex**                                         LAGASQ@IRIT.FR
*IRIT-UPS, 118 route de Narbonne*
*31062 Toulouse Cedex, FRANCE*

## Abstract

Argumentation is based on the exchange and valuation of interacting arguments, followed by the selection of the most acceptable of them (for example, in order to take a decision, to make a choice). Starting from the framework proposed by Dung in 1995, our purpose is to introduce "graduality" in the selection of the best arguments, *i.e.* to be able to partition the set of the arguments in more than the two usual subsets of "selected" and "non-selected" arguments in order to represent different levels of selection. Our basic idea is that an argument is all the more acceptable if it can be preferred to its attackers. First, we discuss general principles underlying a "gradual" valuation of arguments based on their interactions. Following these principles, we define several valuation models for an abstract argumentation system. Then, we introduce "graduality" in the concept of acceptability of arguments. We propose new acceptability classes and a refinement of existing classes taking advantage of an available "gradual" valuation.

## 1. Introduction

As shown by Dung (1995), argumentation frameworks provide a unifying and powerful tool for the study of several formal systems developed for common-sense reasoning, as well as for giving a semantics to logic programs. Argumentation is based on the exchange and valuation of interacting arguments which support opinions and assertions. It can be applied, among others, in the legal domain, for collective decision support systems or for negotiation support.

The fundamental characteristic of an argumentation system is the interaction between arguments. In particular, a relation of attack may exist between arguments. For example, if the argument takes the form of a logical proof, arguments for a proposition and arguments against this proposition can be advanced. In that case, the attack relation relies on logical inconsistency.

The argumentation process is usually divided in two steps: a *valuation* of the relative strength of the arguments, followed by the *selection* of the most *acceptable* arguments.

In the valuation step, it is usual to distinguish two different types of valuations:

- *intrinsic valuation*: here, the value of an argument is independent of its interactions with the other arguments. This enables to simply express to what extent an argument increases the confidence in the statement it supports (see Pollock, 1992; Krause, Ambler, Elvang, & Fox, 1995; Parsons, 1997; Prakken & Sartor, 1997; Amgoud & Cayrol, 1998; Kohlas, Haenni, & Berzati, 2000; Pollock, 2001).

For example, in the work of Krause et al. (1995), using the following knowledge base, composed of *(formula, probability)* pairs $\{(\phi_1, 0.8), (\phi_2, 0.8), (\phi_3, 0.8), ((\phi_1 \wedge \phi_2 \rightarrow \phi_4), 1), ((\phi_1 \wedge \phi_3 \rightarrow \phi_4), 1)\}$, two arguments can be produced[1]:

- $A_1 = <\{\phi_1, \phi_2, (\phi_1 \wedge \phi_2 \rightarrow \phi_4)\}, \phi_4>$
- and $A_2 = <\{\phi_1, \phi_3, (\phi_1 \wedge \phi_3 \rightarrow \phi_4)\}, \phi_4>$.

Both arguments have the same weight $0.8 \times 0.8 \times 1 = 0.64$, and the formula $\phi_4$ has the weight $0.64 + 0.64 - 0.512 = 0.768$[2].

- *interaction-based valuation*: here the value of an argument depends on its attackers (the arguments attacking it), the attackers of its attackers (the defenders), etc.[3]
  Several approaches have been proposed along this line (see Dung, 1995; Amgoud & Cayrol, 1998; Jakobovits & Vermeir, 1999; Besnard & Hunter, 2001) which differ in the sets of values used. Usually, two values are considered. However, there are very few proposals which use more than two values (three values in Jakobovits & Vermeir, 1999, and an infinity of values in Besnard & Hunter, 2001).
  For example, in the work of Besnard and Hunter (2001), the set of values is the interval of the real line $[0, 1]$. In this case, with the set of arguments[4] $\{A_1, A_2, A_3\}$ and considering that $A_1$ attacks $A_2$ which attacks $A_3$, the value of the argument $A_1$ (resp. $A_2$, $A_3$) is 1 (resp. $\frac{1}{2}$, $\frac{2}{3}$).

Intrinsic valuation and interaction-based valuation have often been used separately, according to the considered applications. Some recent works however consider a combination of both approaches (see Amgoud & Cayrol, 1998; Karacapilidis & Papadias, 2001; Pollock, 2001).

Considering now the selection of the more acceptable arguments, it is usual to distinguish two approaches:

- *individual acceptability*: here, the acceptability of *an argument* depends only on its properties. For example, an argument can be said acceptable if and only if it does not have any attacker (in this case, only the interaction between arguments is considered, see Elvang-Goransson et al., 1993). In the context of an intrinsic valuation, an argument can also be said acceptable if and only if it is "better" than each of its attackers (see Amgoud & Cayrol, 1998).
- *collective acceptability*: in this case, the acceptability of a *set of arguments* is explicitly defined. For example, to be acceptable, a set of arguments may not contain two arguments such that one attacks the other (interactions between arguments are used). Dung's (1995) framework is well suited for this kind of approach but allows only for a binary classification: the argument belongs or does not belong to an acceptable set.

---

1. Here, the arguments are under the form of an "Explanation-Conclusion Pair". This is one possible way to compute arguments (see also Lin & Shoham, 1989; Vreeswijk, 1997; Pollock, 1992; Prakken & Sartor, 1997; Simari & Loui, 1992; Elvang-Goransson, Fox, & Krause, 1993; Kohlas et al., 2000; Amgoud & Cayrol, 2002).
2. Weights being probabilities, the weight of an argument is the probability of the conjunction of the formulae of the argument, and the weight of $\phi_4$ is the probability of the disjunction of $A_1$ and $A_2$.
3. Here, we consider only the interactions corresponding to attacks between arguments. There exist also some other types of interactions (for example, arguments which reinforce other arguments instead of attacking them, see Karacapilidis & Papadias, 2001; Verheij, 2002). For this kind of interaction, graduality has not been considered.
4. Here, the initial knowledge base is useless.

It is clear that except for *intrinsic valuations*, most proposals do not allow for any gradual notion of valuation or acceptability (*i.e.* there is a low number of levels to describe values and the acceptability is usually binary). Our aim is therefore to introduce graduality in these two steps.

However, the processes of valuation and of selection are often linked together. This is the case when the selection is done on the basis of the value of arguments[5] or when the selection defines a binary valuation on arguments. We will therefore:

- first consider and discuss the general principles concerning the definition of a *gradual interaction-based valuation* and then define some valuation models in an abstract argumentation system,
- then, introduce the notion of *graduality in the definition of the acceptability* using the previously defined gradual valuations, but also some more classical mechanisms.

Some graduality has already been introduced in argumentation systems. For instance, in the work of Pollock (2001), degrees of justification for beliefs are computed. Arguments are sequences of conclusive and/or *prima-facie* inferences. Arguments are collected in a graph where a node represents the conclusion of an argument, a support link ties a node to nodes from which it is inferred, and an attack link indicates an attack between nodes. The degree of justification of a belief is computed from the strength of the arguments concluding that belief and the strength of the arguments concluding on an attacker of the belief.

Our work takes place in a more abstract framework since we do not consider any argument structure. Our valuation models are based on interactions between arguments and directly apply to arguments.

We use the framework defined by Dung (1995): a set of arguments and a binary attack relation between arguments. We also use a graphical representation of argumentation systems (see Section 2). The gradualisation of interaction-based valuations will be presented in Section 3. Then, in Section 4, we will consider different mechanisms leading to gradual acceptability, sometimes relying on the gradual valuations defined in Section 3. We will conclude in Section 5.

All the proofs of the properties stated in Sections 3 and 4 will be given in Appendix A.

## 2. Dung's (1995) framework and its graphical representation

We consider the abstract framework introduced by Dung (1995). An *argumentation system* $<\mathcal{A}, \mathcal{R}>$ is a set $\mathcal{A}$ of arguments and a binary relation $\mathcal{R}$ on $\mathcal{A}$ called an *attack relation*: consider $A_i$ and $A_j \in \mathcal{A}$, $A_i \mathcal{R} A_j$ means that $A_i$ attacks $A_j$ or $A_j$ is attacked by $A_i$ (also denoted by $(A_i, A_j) \in \mathcal{R}$).

---

5. For example, using Besnard and Hunter's (2001) valuation, we can decide that all the arguments whose value is $> 0.5$ are selected, because 0.5 is the mean value of the set of values; Another possibility, with different valuations (interaction-based or intrinsic), is to accept an argument when its value is better than the value of each of its attackers.

An argumentation system is *well-founded* if and only if there is no infinite sequence $A_0$, $A_1$, ..., $A_n$, ... such that $\forall i, A_i \in \mathcal{A}$ and $A_{i+1}\mathcal{R}A_i$.

Here, we are not interested in the structure of the arguments and we consider an arbitrary attack relation.

**Notation:** $<\mathcal{A}, \mathcal{R}>$ defines a directed graph $\mathcal{G}$ called the *attack graph*. Consider $A \in \mathcal{A}$, the set $\mathcal{R}^-(A)$ is the set of the arguments attacking $A$[6] and the set $\mathcal{R}^+(A)$ is the set of the arguments attacked by $A$[7].

**Example 1**
*The system $<\mathcal{A} = \{A_1, A_2, A_3, A_4\}, \mathcal{R} = \{(A_2, A_3), (A_4, A_3), (A_1, A_2)\}>$ defines the following graph $\mathcal{G}$ with the root*[8] *$A_3$:*

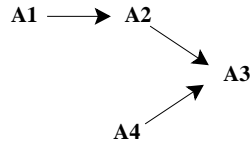

**Definition 1 (Graphical representation of an argumentation system)** *Let $\mathcal{G}$ be the attack graph associated with the argumentation system $<\mathcal{A}, \mathcal{R}>$, we define:*

**Leaf of the attack graph** *A leaf of $\mathcal{G}$ is an argument $A \in \mathcal{A}$ without attackers*[9].

**Path in the attack graph** *A path from $A$ to $B$ is a sequence of arguments $\mathcal{C} = A_1 - \ldots - A_n$ such that:*

- $A = A_1$,
- $A_1 \mathcal{R} A_2$,
- ...,
- $A_{n-1} \mathcal{R} A_n$,
- $A_n = B$.

*The* length of the path *is $n - 1$ (the number of edges that are used in the path) and will be denoted by $l_\mathcal{C}$.*

*A special case is the path*[10] *from $A$ to $A$ whose length is $0$.*

*The set of paths from $A$ to $B$ will be denoted by $\mathcal{C}(A, B)$.*

---

6. $\mathcal{R}^-(A) = \{A_i \in \mathcal{A} | A_i \mathcal{R} A\}$.
7. $\mathcal{R}^+(A) = \{A_i \in \mathcal{A} | A \mathcal{R} A_i\}$.
8. The word "root" is used in an informal sense (it just means that there are in the graph some paths leading to this node). This term and other terms (leaf, branch, path, ...) which are used in this document are standard in graph theory but may have a different definition. They are usual terms in the argumentation domain. Please see Definition 1 in order to know their precise meaning in this document. These definitions simply take into account the fact that the directed edges of our graph link attackers to attacked argument).
9. $A$ is a leaf iff $\mathcal{R}^-(A) = \varnothing$.
10. We will assume that there exists an infinity of such paths. This assumption greatly simplifies the handling of leaves later in the paper.

Graduality in argumentation...

**Dependence, independence, root-dependence of a path**

*Consider 2 paths $\mathcal{C}_A \in \mathcal{C}(A_1, A_n)$ and $\mathcal{C}_B \in \mathcal{C}(B_1, B_m)$.*

*These two paths will be said* dependent *iff $\exists A_i \in \mathcal{C}_A$, $\exists B_j \in \mathcal{C}_B$ such that $A_i = B_j$. Otherwise they are* independent.

*These two paths will be said* root-dependent *in $A_n$ iff $A_n = B_m$ and $\forall A_i \neq A_n \in \mathcal{C}_A$, $\nexists B_j \in \mathcal{C}_B$ such that $A_i = B_j$.*

**Cycles in the attack graph** *A* cycle[11] *is a path $\mathcal{C} = A_1 - \ldots - A_n - A_1$ such that $\forall i, j \in [1, n]$, if $i \neq j$, then $A_i \neq A_j$.*

*A cycle $\mathcal{C}$ is* isolated *iff $\forall A \in \mathcal{C}$, $\nexists B \in \mathcal{A}$ such that $B \mathcal{R} A$ and $B \notin \mathcal{C}$.*

*Two cycles $\mathcal{C}_A = A_1 - \ldots - A_n - A_1$ and $\mathcal{C}_B = B_1 - \ldots - B_m - B_1$ are* interconnected *iff $\exists i \in [1, n], \exists j \in [1, m]$ such that $A_i = B_j$.*

We use the notions of direct and indirect attackers and defenders. The notions introduced here are inspired by related definitions first introduced by Dung (1995) but are not strictly equivalent[12].

**Definition 2 (Direct/Indirect Attackers/Defenders of an argument)** *Consider $A \in \mathcal{A}$:*

- *The* direct attackers *of $A$ are the elements of $\mathcal{R}^-(A)$.*
- *The* direct defenders *of $A$ are the direct attackers of the elements of $\mathcal{R}^-(A)$.*
- *The* indirect attackers *of $A$ are the elements $A_i$ defined by:*
    $\exists \mathcal{C} \in \mathcal{C}(A_i, A)$ *such that $l_\mathcal{C} = 2k + 1$, with $k \geq 1$.*
- *the* indirect defenders *of $A$ are the elements $A_i$ defined by:*
    $\exists \mathcal{C} \in \mathcal{C}(A_i, A)$ *such that $l_\mathcal{C} = 2k$, with $k \geq 2$.*

If the argument $A$ is an attacker (direct or indirect) of the argument $B$, we say that $A$ *attacks $B$* (or that *$B$ is attacked by $A$*). In the same way, if the argument $A$ is a defender (direct or indirect) of the argument $B$, then $A$ *defends $B$* (or *$B$ is defended by $A$*).

Note that an attacker can also be a defender (for example, if $A_1$ attacks $A_2$ which attacks $A_3$, and $A_1$ also attacks $A_3$). In the same way, a direct attacker can be an indirect attacker (for example, if $A_1$ attacks $A_2$ which attacks $A_3$ which attacks $A_4$, and $A_1$ also attacks $A_4$) and the same thing may occur for the defenders.

**Definition 3 (Attack branch and defence branch of an argument)** *Consider $A \in \mathcal{A}$, an* attack branch *(resp.* defence branch*) for $A$ is a path in $\mathcal{G}$ from a leaf to $A$ whose length is odd (resp. even). We say that $A$ is the* root *of an attack branch (resp. a defence branch).*

---

11. This definition of a cycle corresponds to the definition of an elementary cycle in graph theory (an elementary cycle does not contain 2 edges with the same initial extremity, or the same ending extremity).
12. In Dung's (1995) work, direct attackers (resp. defenders) are also indirect attackers (resp. defenders) which is not true in our definitions.

Note that this notion of defence is the basis of the usual notion of reinstatement ($B$ attacks $C$, $A$ attacks $B$ and $C$ is "reinstated" because of $A$). In this paper, reinstatement is taken into account indirectly, because the value of the argument $C$ and the possibility for selecting $C$ will be increased thanks to the presence of $A$.

All these notions are illustrated on the following example:

**Example 2**

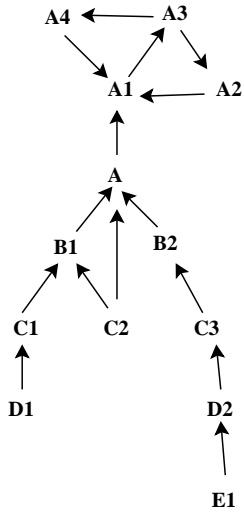

On this graph $\mathcal{G}$, we can see:

- a path from $C_2$ to $A$ whose length is 2 ($C_2 - B_1 - A$),
- 2 cycles $A_1 - A_3 - A_2 - A_1$ and $A_1 - A_3 - A_4 - A_1$, of length 3, which are not isolated (note that $A_1 - A_3 - A_2 - A_1 - A_3 - A_4 - A_1$ is not a cycle with our definition),
- the two previous cycles are interconnected (in $A_1$ and $A_3$),
- the paths $D_1 - C_1 - B_1$ and $C_3 - B_2 - A$ are independent, the paths $D_1 - C_1 - B_1 - A$ and $C_3 - B_2 - A$ are root-dependent and the paths $D_1 - C_1 - B_1 - A$ and $C_2 - B_1 - A$ are dependent,
- $D_1$, $C_2$, $E_1$ are the leaves of $\mathcal{G}$,
- $D_1 - C_1 - B_1 - A$ is an attack branch for $A$ whose length is 3, $C_2 - B_1 - A$ is a defence branch for $A$ whose length is 2,
- $C_2$, $B_1$ and $B_2$ are the direct attackers of $A$,
- $C_1$, $C_2$ (which is already a direct attacker of $A$) and $C_3$ are the direct defenders of $A$,
- $D_1$ and $D_2$ are the two indirect attackers of $A$,
- $E_1$ is the only indirect defender of $A$.

## 3. Graduality in interaction-based valuations

We consider two different valuation methods for taking into account the quality of attackers and defenders of an argument in order to define the value of an argument using only the interaction between arguments[13]:

- In the first approach, the value of an argument only depends on the values of the direct attackers of this argument. Therefore, defenders are taken into account through the attackers. This approach is called *local*.
- In the second approach, the value of an argument represents the set of all the attack and the defence branches for this argument. This approach is called *global*.

The main difference between these two approaches is illustrated by the following example:

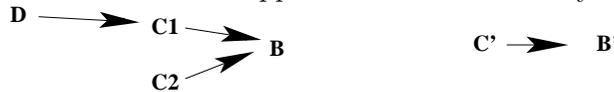

---

13. We pursue a work initiated in (Cayrol & Lagasquie-Schiex, 2003c) and propose some improvements.

In the local approach, $B$ has two direct attackers ($C_2$ and $C_1$) whereas $B'$ has only one ($C'$). Thus $B'$ is better than $B$ (since $B'$ suffers one attack whereas $B$ suffers two attacks). In the global approach, two branches (one of attack and one of defence) lead to $B$ whereas only one branch of attack leads to $B'$. Thus $B$ is better than $B'$ (since it has at least one defence whereas $B'$ has none). In this case, $C_1$ loses its negative status of attacker, since it is in fact "carrying a defence" for $B$.

### 3.1 Local approach (generic valuation)

Some existing proposals can already be considered as examples of *local valuations*.

In Jakobovits and Vermeir's (1999) approach, a labelling of a set of arguments assigns a status (accepted, rejected, undecided) to each argument using labels from the set $\{+, -, ?\}$. $+$ (resp. $-$, ?) represents the "accepted" (resp. "rejected", "undecided") status. Intuitively, an argument labelled with ? is both supported and weakened.

**Definition 4 (Jakobovits and Vermeir's labellings, 1999)** *Let $<\mathcal{A}, \mathcal{R}>$ be an argumentation system. A complete labelling of $<\mathcal{A}, \mathcal{R}>$ is a function $Lab : \mathcal{A} \to \{+, ?, -\}$ such that:*

1. *If $Lab(A) \in \{?, -\}$ then $\exists B \in \mathcal{R}^-(A)$ such that $Lab(B) \in \{+, ?\}$*
2. *If $Lab(A) \in \{+, ?\}$ then $\forall B \in \mathcal{R}^-(A) \cup \mathcal{R}^+(A)$, $Lab(B) \in \{?, -\}$*

The underlying intuition is that an argument can only be weakened (label $-$ or ?) if one of its direct attackers is supported (condition 1); an argument can get a support only if all its direct attackers are weakened and an argument which is supported (label $+$ or ?) weakens the arguments it attacks (condition 2). So:

- If $A$ has no attacker $Lab(A) = +$.
- If $Lab(A) = ?$ then $\exists B \in \mathcal{R}^-(A)$ such that $Lab(B) = ?$.
- If $(\forall B \in \mathcal{R}^-(A), Lab(B) = -)$ then $Lab(A) = +$.
- If $Lab(A) = +$ then $\forall B \in \mathcal{R}^-(A) \cup \mathcal{R}^+(A)$, $Lab(B) = -$.

Every argumentation system can be completely labelled. The associated semantics is that $S$ is an acceptable set of arguments iff there exists a complete labelling $Lab$ of $<\mathcal{A}, \mathcal{R}>$ such that $S = \{A | Lab(A) = +\}$.
Other types of labellings are introduced by Jakobovits and Vermeir (1999) among which the so-called "rooted labelling" which induces a corresponding "rooted" semantics. The idea is to reject only the arguments attacked by accepted arguments: an attack by an "undecided" argument is not rooted since an "undecided" attacker may become rejected.

**Definition 5 (Jakobovits and Vermeir's labellings, 1999 – continuation)**
*The complete labelling Lab is rooted iff $\forall A \in \mathcal{A}$, if $Lab(A) = -$ then $\exists B \in \mathcal{R}^-(A)$ such that $Lab(B) = +$.*

The rooted semantics enables to clarify the links between all the other semantics introduced by Jakobovits and Vermeir (1999) and some semantics introduced by Dung (1995).

**Example 3** *On the following example:*

$$A_n \longrightarrow A_{n-1} \dashrightarrow A_2 \longrightarrow A_1$$

*For $n$ even, we obtain $Lab(A_n) = Lab(A_{n-2}) = \ldots = Lab(A_2) = +$ and $Lab(A_{n-1}) = Lab(A_{n-3}) = \ldots = Lab(A_1) = -$.*
*For $n$ odd, we obtain $Lab(A_n) = Lab(A_{n-2}) = \ldots = Lab(A_1) = +$ and $Lab(A_{n-1}) = Lab(A_{n-3}) = \ldots = Lab(A_2) = -$*

Another type of *local valuation* has been introduced recently by Besnard and Hunter (2001) for "deductive" arguments. The approach can be characterised as follows. An argument is structured as a pair $\langle support, conclusion \rangle$, where *support* is a consistent set of formulae that enables to prove the formula *conclusion*. The attack relation considered here is strict and cycles are not allowed. The notion of a "tree of arguments" allows a concise and exhaustive representation of attackers and defenders of a given argument, root of the tree. A function, called a "categoriser", assigns a value to a tree of arguments. This value represents the relative strength of an argument (root of the tree) given all its attackers and defenders. Another function, called an "accumulator", synthesises the values assigned to all the argument trees whose root is an argument for (resp. against) a given conclusion. The phase of categorisation therefore corresponds to an interaction-based valuation. Besnard and Hunter (2001) introduce the following function $Cat$:

- if $\mathcal{R}^-(A) = \varnothing$, then $Cat(A) = 1$
- if $\mathcal{R}^-(A) \neq \varnothing$ with $\mathcal{R}^-(A) = \{A_1, \ldots, A_n\}$, $Cat(A) = \frac{1}{1+Cat(A_1)+\ldots+Cat(A_n)}$

Intuitively, the larger the number of direct attackers of an argument, the lower its value. The larger the number of defenders of an argument, the larger its value.

**Example 3 (continuation)** *We obtain:*
*$Cat(A_n) = 1$, $Cat(A_{n-1}) = 0.5$, $Cat(A_{n-2}) = 0.66$, $Cat(A_{n-3}) = 0.6$, ..., and $Cat(A_1) = (\sqrt{5}-1)/2$ when $n \to \infty$ (this value is the inverse of the golden ratio[14]).*
*So, we have:*
*If $n$ is even $Cat(A_{n-1}) \leq \ldots \leq Cat(A_3) \leq Cat(A_1) \leq Cat(A_2) \leq \ldots \leq Cat(A_n) = 1$*
*If $n$ is odd $Cat(A_{n-1}) \leq \ldots \leq Cat(A_2) \leq Cat(A_1) \leq Cat(A_3) \leq \ldots \leq Cat(A_n) = 1$*

Our approach for *local valuations* is a generalisation of these two previous proposals in the sense that Besnard and Hunter's (2001) $Cat$ function and Jakobovits and Vermeir's (1999) labellings are instances of our approach.
The main idea is that the value of an argument is obtained with the composition of two functions:

- one for aggregating the values of all the direct attackers of the argument; so, this function computes the value of the "direct attack";
- the other for computing the effect of the "direct attack" on the value of the argument: if the value of the "direct attack" increases then the value of this argument decreases, if the value of the "direct attack" decreases then the value of this argument increases.

---

14. The golden ratio is a famous number since the antiquity which has several interesting properties in several domains (architecture, for example).

Let $(W, \geq)$ be a totally ordered set with a minimum element ($V_{\text{Min}}$) and a subset $V$ of $W$, that contains $V_{\text{Min}}$ and with a maximum element $V_{\text{Max}}$.

**Definition 6 (Generic gradual valuation)** *Let $<\mathcal{A}, \mathcal{R}>$ be an argumentation system. A valuation is a function $v : \mathcal{A} \to V$ such that:*

1. $\forall A \in \mathcal{A}, v(A) \geq V_{Min}$
2. $\forall A \in \mathcal{A}$, if $\mathcal{R}^-(A) = \emptyset$, then $v(A) = V_{Max}$
3. $\forall A \in \mathcal{A}$, if $\mathcal{R}^-(A) = \{A_1, \ldots, A_n\} \neq \emptyset$, then $v(A) = g(h(v(A_1), \ldots, v(A_n)))$

*with $h : V^* \to W$ such that ($V^*$ denotes the set of all finite sequences of elements of $V$)*

- $h(x) = x$
- $h() = V_{Min}$
- *For any permutation $(x_{i1}, \ldots, x_{in})$ of $(x_1, \ldots, x_n)$, $h(x_{i1}, \ldots, x_{in}) = h(x_1, \ldots, x_n)$*
- $h(x_1, \ldots, x_n, x_{n+1}) \geq h(x_1, \ldots, x_n)$
- *if $x_i \geq x'_i$ then $h(x_1, \ldots, x_i, \ldots, x_n) \geq h(x_1, \ldots, x'_i, \ldots, x_n)$*

*and $g : W \to V$ such that*

- $g(V_{Min}) = V_{Max}$
- $g(V_{Max}) < V_{Max}$
- *$g$ is non-increasing (if $x \leq y$ then $g(x) \geq g(y)$)*

Note that $h(x_1, \ldots, x_n) \geq \max(x_1, \ldots, x_n)$ is a logical consequence of the properties of the function $h$.

A first property on the function $g$ explains the behaviour of the local valuation in the case of an argument which is the root of only one branch (like in Example 3):

**Property 1** *The function $g$ satisfies for all $n \geq 1$:*

$$g(V_{Max}) \leq g^3(V_{Max}) \leq \ldots \leq g^{2n+1}(V_{Max}) \leq g^{2n}(V_{Max}) \leq \ldots \leq g^2(V_{Max}) \leq V_{Max}$$

*Moreover, if $g$ is strictly non-increasing and $g(V_{Max}) > V_{Min}$, the previous inequalities become strict.*

A second property shows that the local valuation induces an ordering relation on arguments:

**Property 2 (Complete preordering)** *Let $v$ be a valuation in the sense of Definition 6. $v$ induces a complete[15] preordering $\succeq$ on the set of arguments $\mathcal{A}$ defined by: $A \succeq B$ iff $v(A) \geq v(B)$.*

A third property handles the cycles:

---

15. A complete preordering on $\mathcal{A}$ means that any two elements of $\mathcal{A}$ are comparable.

**Property 3 (Value in a cycle)** *Let $\mathcal{C}$ be an isolated cycle of the attack graph, whose length is n. If n is odd, all the arguments of the cycle have the same value and this value is a fixpoint of the function g. If n is even, the value of each argument of the cycle is a fixpoint of the function $g^n$.*

The following property shows the underlying principles satisfied by all the local valuations defined according to our schema:

**Property 4 (Underlying principles)** *The gradual valuation given by Definition 6 respects the following principles:*

**P1** *The valuation is maximal for an argument without attackers and non maximal for an attacked and undefended argument.*

**P2** *The valuation of an argument is a function of the valuation of its direct attackers (the "direct attack").*

**P3** *The valuation of an argument is a non-increasing function of the valuation of the "direct attack".*

**P4** *Each attacker of an argument contributes to the increase of the valuation of the "direct attack" for this argument.*

The last properties explain why Jakobovits and Vermeir (1999) and Besnard and Hunter (2001) propose instances of the local valuation described in Definition 6:

**Property 5 (Link with Jakobovits & Vermeir, 1999)**
*Every rooted labelling of $<\mathcal{A}, \mathcal{R}>$ in the sense of Jakobovits and Vermeir (1999) can be defined as an instance of the generic valuation such that:*

- $V = W = \{-, ?, +\}$ *with* $- < ? < +$,
- $V_{Min} = -$,
- $V_{Max} = +$,
- *g defined by* $g(-) = +$, $g(+) = -$, $g(?) = ?$
- *and h is the function* max.

**Property 6 (Link with Besnard & Hunter, 2001)** *The gradual valuation of Besnard and Hunter (2001) can be defined as an instance of the generic valuation such that:*

- $V = [0, 1]$,
- $W = [0, \infty[$,
- $V_{Min} = 0$,
- $V_{Max} = 1$,
- $g : W \to V$ *defined by* $g(x) = \frac{1}{1+x}$
- *and h defined by* $h(x_1, \ldots, x_n) = x_1 + \ldots + x_n$.

Note that, in the work of Besnard and Hunter (2001), the valued graphs are acyclic. However, it is easy to show that the valuation proposed by Besnard and Hunter (2001) can be generalised to graphs with cycles (in this case, we must solve second degree equations – see Example 5).

**Example 4** *Consider the following graph:*

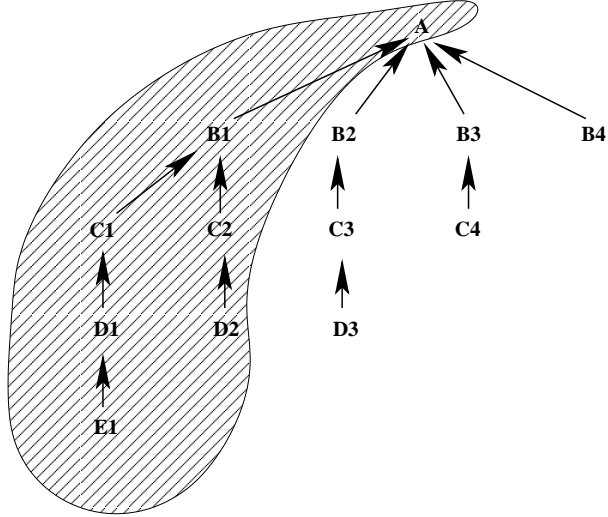

*In this example, with the generic valuation, we obtain:*

- $v(E_1) = v(D_2) = v(D_3) = v(C_4) = v(B_4) = V_{Max}$
- $v(D_1) = v(C_2) = v(C_3) = v(B_3) = g(V_{Max})$
- $v(C_1) = v(B_2) = g^2(V_{Max})$
- $v(B_1) = g(h(g^2(V_{Max}), g(V_{Max})))$
- $v(A) = g(h(g(h(g^2(V_{Max}), g(V_{Max}))), g^2(V_{Max}), g(V_{Max}), V_{Max}))$

*So, we have:*

$$E_1, D_2, D_3, C_4, B_4$$
$$\succeq$$
$$C_1, B_2$$
$$\succeq$$
$$D_1, C_2, C_3, B_3$$

*However, the constraints on $v(A)$ and $v(B_1)$ are insufficient to compare $A$ and $B_1$ with the other arguments.*

*The same problem exists if we reduce the example to the hatched part of the graph in the previous figure; we obtain $E_1, D_2 \succeq C_1 \succeq D_1, C_2$, but $A$ and $B_1$ cannot be compared with the other arguments*[16].

*Now, we use the instance of the generic valuation proposed by Besnard and Hunter (2001):*

- $v(E_1) = v(D_2) = v(D_3) = v(C_4) = v(B_4) = 1$,
- $v(D_1) = v(C_2) = v(C_3) = v(B_3) = \frac{1}{2}$,
- $v(C_1) = v(B_2) = \frac{2}{3}$,

---

16. $v(A) = g^2(h(g^2(V_{Max}), g(V_{Max})))$ and $v(B_1) = g(h(g^2(V_{Max}), g(V_{Max})))$.

- $v(B_1) = \frac{6}{13}$,
- $v(A) = \frac{78}{283}$.

So, we have:

$$E_1, D_2, D_3, C_4, B_4$$
$$\succeq$$
$$C_1, B_2$$
$$\succeq$$
$$D_1, C_2, C_3, B_3$$
$$\succeq$$
$$B_1$$
$$\succeq$$
$$A$$

*However, if we reduce the example to the hatched part of the graph, then the value of $A$ is $\frac{13}{19}$. So, $v(A)$ is better than $v(B_1)$ and $v(D_1)$, but also than $v(C_1)$ ($A$ becomes better than its defender).*

**Example 5 (Isolated cycle)** *Consider the following graph reduced to an isolated cycle:*

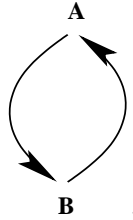

*A generic valuation gives $v(A) = v(B) =$ fixpoint of $g^2$.*
*If we use the instance proposed by Besnard and Hunter (2001), $v(A)$ and $v(B)$ are solutions of the following second degree equation: $x^2 + x - 1 = 0$.*
*So, we obtain: $v(A) = v(B) = \frac{-1+\sqrt{5}}{2} \approx 0.618$ (the inverse of the golden ratio again).*

### 3.2 Global approach (with tuples)

We now consider a second approach for the valuation step, called the global approach. Here, the key idea is that the value of $A$ must describe the subgraph whose root is $A$. So, we want to memorise the length of each branch leading to $A$ in a tuple (for an attack branch, we have an odd integer, and for a defence branch, we have an even integer).

In this approach, the main constraint is that we must be able to identify the branches leading to the argument and to compute their lengths. This is very easy in the case of an acyclic graph. We therefore introduce first a global gradual valuation for acyclic graphs. Then, in the next sections, we extend our proposition to the case of graphs with cycles, and we study the properties of this global gradual valuation.

#### 3.2.1 Gradual valuation with tuples for acyclic graphs

First, in order to record the lengths of the branches leading to the arguments, we use the notion of tuples and we define some operations on these tuples:

**Definition 7 (Tuple)** *A tuple is a sequence of integers. The tuple* $\underbrace{(0,\ldots,0,\ldots)}_{\infty}$ *will be denoted by* $0^\infty$. *The tuple* $\underbrace{(1,\ldots,1,\ldots)}_{\infty}$ *will be denoted by* $1^\infty$.

**Notation 1** $\mathcal{T}$ *denotes the set of the tuples built with positive integers.*

**Definition 8 (Operations on the tuples)** *We have two kinds of operations on tuples:*

- *the concatenation of two tuples is defined by the function* $\star : \mathcal{T} \times \mathcal{T} \to \mathcal{T}$ *such that*

$$0^\infty \star t = t \star 0^\infty = t \text{ for } t \neq ()$$
$$(x_1,\ldots,x_n,\ldots) \star (x'_1,\ldots,x'_n,\ldots) = \texttt{Sort}(x_1,\ldots,x_n,\ldots,x'_1,\ldots,x'_n,\ldots)$$

  $\texttt{Sort}$ *being the function which orders a tuple by increasing values.*

- *the addition of a tuple and an integer is defined by the function* $\oplus : \mathcal{T} \times \mathbb{N} \to \mathcal{T}$ *such that*

$$0^\infty \oplus k = (k)$$
$$() \oplus k = ()$$
$$(x_1,\ldots,x_n) \oplus k = (x_1+k,\ldots,x_n+k)$$
$$(x_1,\ldots,x_n,\ldots) \oplus k = (x_1+k,\ldots,x_n+k,\ldots) \text{ if } (x_1,\ldots,x_n,\ldots) \neq 0^\infty$$

Note that we allow infinite tuples, among other reasons, because they are needed later in order to compute the ordering relations described in Section 3.2.4 (in particular when the graph is cyclic).

The operations on the tuples have the following properties:

**Property 7 (Properties of $\star$ and $\oplus$)**
*The concatenation $\star$ is commutative and associative.*
*For any tuple $t$ and any integers $k$ and $k'$, $(t \oplus k) \oplus k' = t \oplus (k + k')$.*
*For any integer $k$ and any tuples $t$ and $t'$ different from $0^\infty$[17], $(t \star t') \oplus k = (t \oplus k) \star (t' \oplus k)$.*

In order to valuate the arguments, we split the set of the lengths of the branches leading to the argument in two subsets, one for the lengths of defence branches (even integers) and the other one for the lengths of attack branches (odd integers). This is captured by the notion of tupled values:

**Definition 9 (Tupled value)** *A tupled value is a pair of tuples $vt = [vt_p, vt_i]$ with:*

- $vt_p$ *is a tuple of even integers ordered by increased values; this tuple is called the even component of $vt$;*
- $vt_i$ *is a tuple of odd integers ordered by increased values; this tuple is called the odd component of $vt$.*

---

17. Otherwise it is false : $(0^\infty \star (p)) \oplus k = (p+k)$, whereas $(0^\infty \oplus k) \star ((p) \oplus k) = (k) \star (p+k) = (k, p+k)$.

**Notation 2** $\mathcal{V}$ denotes the subset of $\mathcal{T} \times \mathcal{T}$ of all tupled values (so, $\forall vt \in \mathcal{V}$, $vt$ is a pair of tuples satisfying Definition 9).

Using this notion of tupled-values, we can define the *computation process* of the gradual valuation with tuples[18] in the case of acyclic graphs.

**Definition 10 (Valuation with tuples for acyclic graphs)** Let $<\mathcal{A}, \mathcal{R}>$ be an argumentation system without cycles. A valuation with tuples is a function $v : \mathcal{A} \to \mathcal{V}$ such that:

If $A \in \mathcal{A}$ is a leaf then
$$v(A) = [0^\infty, ()].$$

If $A \in \mathcal{A}$ has direct attackers denoted by $B_1, \ldots, B_n, \ldots$ then

$$v(A) = [v_p(A), v_i(A)] \text{ with: } \begin{vmatrix} v_p(A) = (v_i(B_1) \oplus 1) \star \ldots \star (v_i(B_n) \oplus 1) \star \ldots \\ v_i(A) = (v_p(B_1) \oplus 1) \star \ldots \star (v_p(B_n) \oplus 1) \star \ldots \end{vmatrix}$$

**Notes:** The choice of the value $[0^\infty, ()]$ for the leaves is justified by the fact that the value of an argument memorises all the lengths of the branches leading to the argument. Using the same constraint, either $v_p(A)$ or $v_i(A)$ may be empty but not both[19].
Note also that the set of the direct attackers of an argument can be infinite (this property will be used when we take into account an argumentation graph with cycles).

**Example 6** *On this graph, the valuation with tuples gives the following results:*

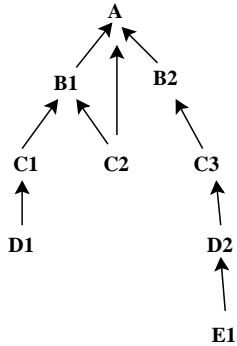

*On this graph $\mathcal{G}$, we have:*

- $v(D_1) = v(C_2) = v(E_1) = [0^\infty, ()]$,
- $v(C_1) = v(D_2) = [(), (1)]$,
- $v(C_3) = [(2), ()]$,
- $v(B_1) = [(2), (1)]$,
- $v(B_2) = [(), (3)]$,
- $v(A) = [(2, 4), (1, 3)]$.

---

18. This definition is different from the definition given in (Cayrol & Lagasquie-Schiex, 2003c). The ideas are the same but the formalisation is different.
19. The proof is the following:.

- If $A$ is not a leaf, at least one of the tuples is not empty, because there exists at least one branch whose length is $> 0$ leading to $A$ (see Definitions 8 and 10).
- And, if $A$ is a leaf, there also exists at least one defence branch because the path from $A$ to $A$ is allowed and its length is 0 (in fact, there are an infinity of such paths – see Definition 1) and no attack branch leading to the leaf (see Definition 10).

So, the value of a leaf is $[0^\infty, ()]$, and it is impossible that $v_p(A) = v_i(A) = ()$.

### 3.2.2 Study of cycles

Handling cycles raises some important issues: the notion of branch is not always useful in a cycle (for example, in an unattacked cycle like in Examples 5 and 7), and when this notion is useful, the length of a branch can be defined in different ways.
Let us consider different examples:

**Example 7 (Unattacked cycle)** *The graph is reduced to an unattacked cycle $A - B - A$ which attacks the argument C:*

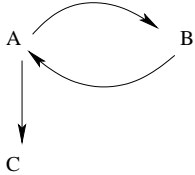

*The notion of branch is useless in this case, because there is no leaf in the graph.*
*There are two possibilities:*

- *First, one can consider that the cycle is like an infinite branch; so A (resp. B) is the root of one branch whose length is $\infty$. But the parity of the length of this branch is undefined, and it is impossible to say if this branch is an attack branch or a defence branch.*
- *The second possibility is to consider that the cycle is like an infinity of branches; so A (resp. B) is the root of an infinity of attack branches and defence branches whose lengths are known and finite.*

*The second possibility means that the cycle may have two representations which are acyclic but also infinite graphs (one with the root A and the other one with the root B). This is a rewriting process of the cycle:*

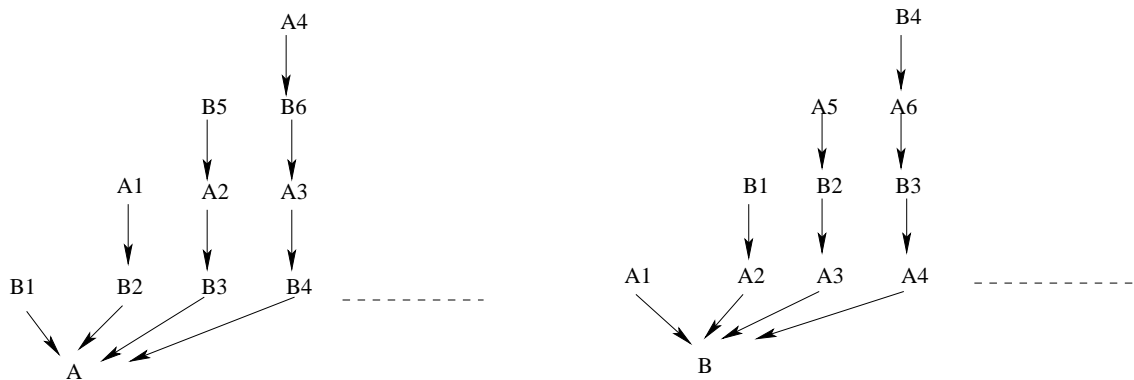

*The $A_i$ and $B_i$ must be new arguments created during the rewriting process of the cycle.*

**Example 8 (Attacked cycle)** *The cycle $A - B - A$ is attacked by at least one argument which does not belong to the cycle (here, the attacker is the unattacked argument D):*

259

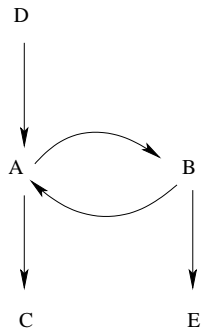

*In this case, the notion of branch is useful because there exists one leaf in the graph, but the difficulty is to compute the length of this branch. As in Example 7, we can consider either that there is only one infinite branch (so, it is impossible to know if this branch is an attack or a defence branch), or that there is an infinity of attack branches and defence branches whose lengths are known and finite.*
*In the second case, the graph can be rewritten into the following structures:*

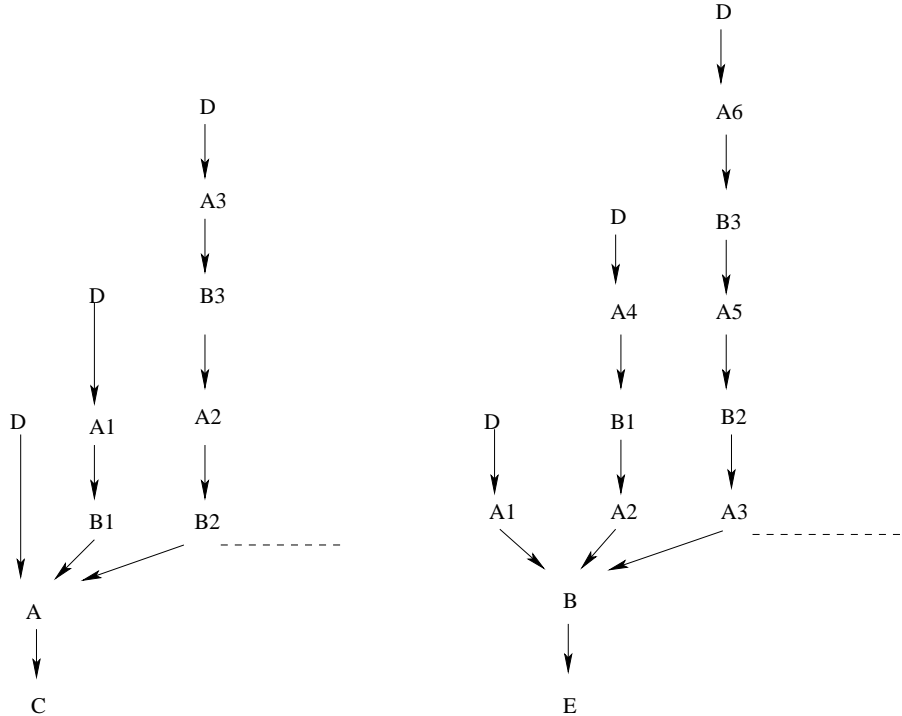

*The $A_i$ and $B_i$ must be new arguments created during the rewriting process of the graph.*

From the previous examples, we have chosen to manage a cycle as an infinity of attack branches and defence branches whose lengths are known and finite because we would like to be able to apply Definition 10 in all cases (acyclic graphs and graphs with cycles). However, we need a rewriting process of the graph with cycles into an acyclic graph. There are two different cases, one for the unattacked cycles and one for the attacked cycles:

**Definition 11 (Rewriting of an unattacked cycle)** *Let $\mathcal{C} = A_0 - A_1 - \ldots - A_{n-1} - A_0$ an unattacked cycle. The graph $\mathcal{G}$ which contains $\mathcal{C}$ is rewritten as follows:*

1. the cycle $\mathcal{C}$ is removed,

2. and replaced by the infinite acyclic graphs, one for each $A_i$, $i = 0 \ldots n-1$:

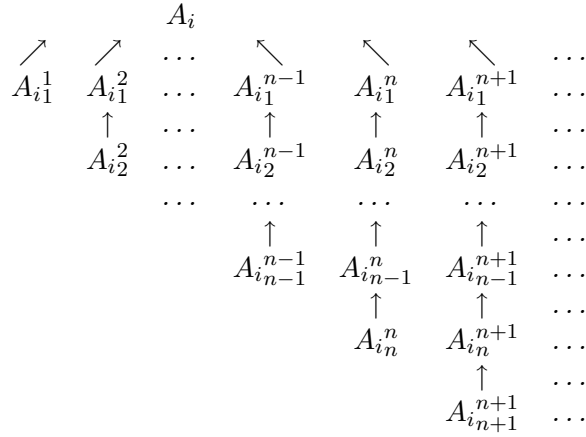

3. the edges between each of the $A_i$ and an argument which does not belong to $\mathcal{C}$ are kept.

**Example 7 – Unattacked cycle (continuation)** The graph $\mathcal{G}$ containing the unattacked cycle $A - B - A$ and the argument $C$, which is attacked by $A$, is rewritten as follows:

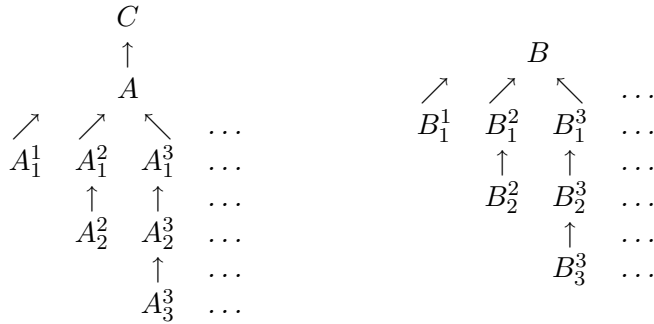

where the $A_k^l$ and $B_k^l$ are new arguments.

**Definition 12 (Rewriting of an attacked cycle)** Let $\mathcal{C} = A_0 - A_1 - \ldots - A_{n-1} - A_0$ an attacked cycle, the direct attacker of each $A_i$ is denoted $B_i$, if it exists. The graph $\mathcal{G}$ which contains $\mathcal{C}$ is rewritten as follows:

1. the cycle $\mathcal{C}$ is removed,

2. and replaced by the infinite acyclic graphs, one for each $A_i$ $i = 0 \ldots n-1$:

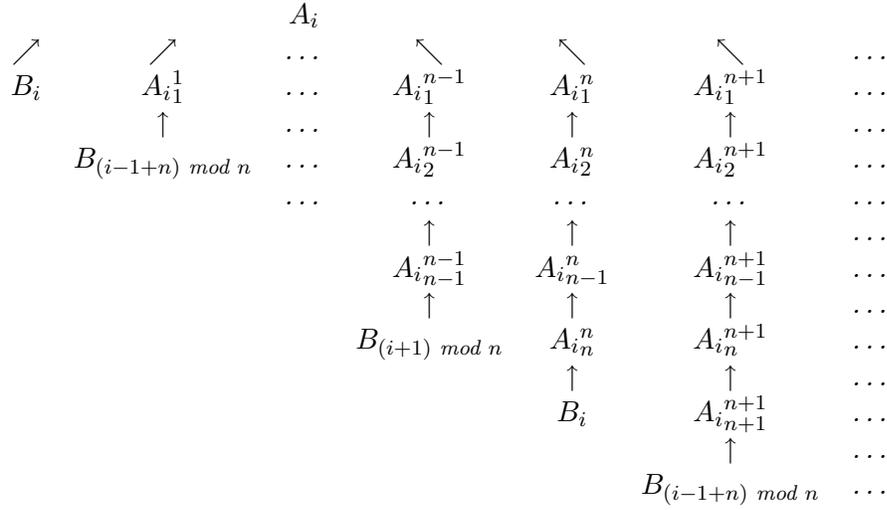

(the branches leading to $B_k$ exist iff $B_k$ exists[20]).

3. the edges between each of the $A_i$ and an argument which does not belong to $\mathcal{C}$ are kept.

4. the edges between each of the $B_i$ and an argument which does not belong to $\mathcal{C}$ are kept.

**Example 8 – Attacked cycle (continuation)** The graph $\mathcal{G}$ containing the cycle $A - B - A$ attacked in $A$ by the argument $D$ and with the argument $C$ (resp. $E$) attacked by $A$ (resp. $B$) is rewritten as follows:

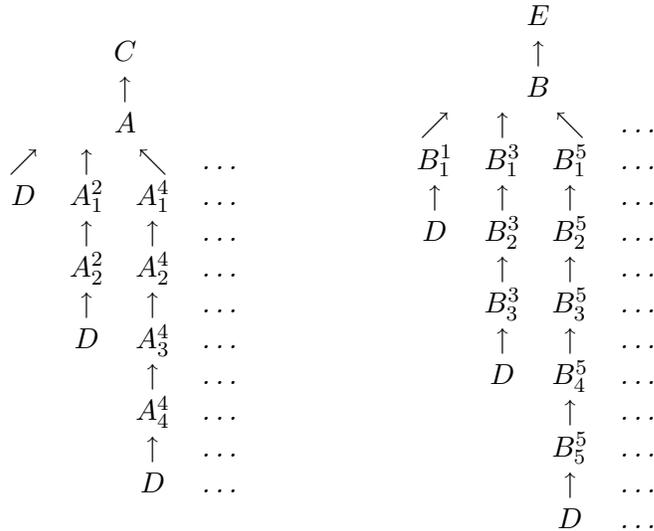

where the $A_k^l$ and $B_k^l$ are new arguments.

---

20. The operator mod is the modulo function.

**Note:** If there exist several cycles in a graph, we have two cases.

- If they are not interconnected, we rewrite each cycle, and the valuation of the resulting graph after rewriting does not depend on the order of cycles we select to rewrite because the valuation process only uses the length of the branches.

- If they are interconnected, they are considered as a metacyle which is in turn attacked or unattacked and the previous methodology can be used leading to a more complex rewriting process which is not formalized here (see details and examples in Appendix B).

### 3.2.3 A gradual valuation with tuples for general graphs

Using the definitions given in Sections 3.2.1 and 3.2.2, the gradual valuation with tuples given by Definition 10 is applicable for arbitrary graphs *after the rewriting process*.
Let us apply the rewriting process and Definition 10 on different examples.

**Example 7 – Unattacked cycle (continuation)**
*Consider the following graph:*

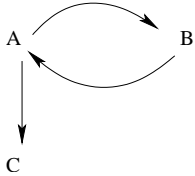

*The rewriting of this graph has been given in Section 3.2.2.*

*Definition 10 produces:*

$$v_p(A) = (v_i(A_1^1) \oplus 1) \star \ldots \star (v_i(A_1^n) \oplus 1) \star \ldots$$

$$v_i(A) = (v_p(A_1^1) \oplus 1) \star \ldots \star (v_p(A_1^n) \oplus 1) \star \ldots$$

*Applying Definition 10 for different arguments in the rewritten graph produces the following equalities:*

- $v(A_n^n) = [0^\infty, ()]$ *for each* $n \geq 1$
- $v(A_{n-1}^n) = [(), (1)]$ *for each* $n \geq 2$
- $v(A_n^m) = [v_p(A_{n+2}^m) \oplus 2, v_i(A_{n+2}^m) \oplus 2]$ *for each* $n \geq 1$ *and* $m \geq n+2$

*So, using the above equalities in the formulae giving $v_p(A)$ and $v_i(A)$, we define two sequences of tuples : a sequence $(x_k, k \geq 1)$ of infinite tuples of even integers, and a sequence $(y_k, k \geq 1)$ of infinite tuples of odd integers*

$$x_k = (2) \star (v_i(A_{2k-1}^{2k+1}) \oplus 1) \star \ldots \star (v_i(A_{2k-1}^n) \oplus 1) \star \ldots$$

$$y_k = (1) \star (v_p(A_{2k-1}^{2k+1}) \oplus 1) \star \ldots \star (v_p(A_{2k-1}^n) \oplus 1) \star \ldots$$

*From the results stated in Property 7, it is easy to prove that $v_p(A) = x_1$ and for each $k \geq 1$, $x_k = (2) \star (x_{k+1} \oplus 2)$.*

*Similarly, $v_i(A) = y_1$ and for each $k \geq 1$, $y_k = (1) \star (y_{k+1} \oplus 2)$.*

*These equations enable to prove that :*

    *For each even integer $p$ $p > 0$, $p$ belongs to each tuple $x_i, i \geq 1$.*

    *For each odd integer $p$, $p$ belongs to each tuple $y_i$, $i \geq 1$.*

*The proof is done by induction on $p$.*

*So, $v(A) = v(B) = [(2, 4, 6, \ldots), (1, 3, 5, \ldots)]$.*
*Then, $v(C) = [(2, 4, 6, \ldots), (3, 5, 7, \ldots)]$.*

*Note that all the above results can be readily extended to an unattacked cycle of length n, $n \geq 2$.*

**Property 8 (Properties of unattacked cycles)**
*For each unattacked cycle, for each argument $A$ of the cycle, $v(A) = [(2, 4, 6, \ldots), (1, 3, 5, \ldots)]$.*

**Example 8 – Attacked cycle (continuation)**    *Consider the following graph:*

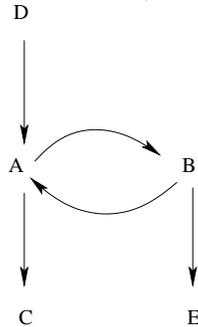

*The rewriting of this graph has been given in Section 3.2.2.*

*Definition 10 produces:*

$$v_p(A) = (v_i(D) \oplus 1) \star (v_i(A_1^2) \oplus 1) \star \ldots \star (v_i(A_1^{2n}) \oplus 1) \star \ldots$$

$$v_i(A) = (v_p(D) \oplus 1) \star (v_p(A_1^2) \oplus 1) \star \ldots \star (v_p(A_1^{2n}) \oplus 1) \star \ldots$$

*and also*

$v(D) = [0^\infty, ()]$

$v(A_n^n) = [(), (1)]$ *for each $n \geq 2$*

*As done in the treatment of Example 7, the formulae giving $v_p(A)$ and $v_i(A)$ can be rewritten in order to bring to light some interesting sequences of tuples.*

$$x'_k = (v_i(A^{2k}_{2k-1}) \oplus 1) \star \ldots \star (v_i(A^{2(k+p)}_{2k-1}) \oplus 1) \star \ldots$$

$$y'_k = (1) \star (v_p(A^{2k}_{2k-1}) \oplus 1) \star \ldots \star (v_p(A^{2(k+p)}_{2k-1}) \oplus 1) \star \ldots$$

Then, it is easy to prove that $v_p(A) = x'_1$ and for each $k \geq 1$, $x'_k = (x'_{k+1} \oplus 2)$.

Similarly, $v_i(A) = y'_1$ and for each $k \geq 1$, $y'_k = (1) \star (y'_{k+1} \oplus 2)$.

The first equation enables to prove that $x'_1$ is the empty tuple[21].

The second equation has already been solved and produces $y'_1 = (1, 3, 5, \ldots)$.

So, $v(A) = [(), (1, 3, 5, \ldots)]$. For $B$, we can reason as for $A$, and we have $v(B) = [(2, 4, 6, \ldots), ()]$. Then, $v(C) = [(2, 4, 6, \ldots), ()]$, $v(E) = [(), (3, 5, 7 \ldots)]$.

**Notation:** in order to simplify the writing, we will not repeat the values inside the tuples (we will just indicate under each value how many times it appears). For example:

$$[(2, 4, 4, 6, 6, 6, 8, 8, 8, 8 \ldots), (3, 5, 5, 7, 7, 7, 9, 9, 9, 9 \ldots)]$$

will be denoted by

$$[(2, \underbrace{4}_{2}, \underbrace{6}_{3}, \underbrace{8}_{4}, \ldots), (3, \underbrace{5}_{2}, \underbrace{7}_{3}, \underbrace{9}_{4}, \ldots)]$$

**Conclusion about cycles** Cycles are expensive since all the values obtained are infinite. In appendix B, we introduce an algorithm for computing these tupled values. It uses a process of value propagation and is parameterised by a maximum "number of runs through a cycle". This number will be used in order to stop the propagation mechanism and to obtain finite (thus incomplete) tupled values.

3.2.4 Comparison of tupled values

In this section, we define the comparison relation between arguments (so, between some particular tupled values), using the following idea: an argument $A$ is better than an argument $B$ iff $A$ has a better defence (for it) and a lower attack (against it).

The first idea is to use a lexicographic ordering on the tuples. This lexicographic ordering denoted by $\leq_{lex\infty}$ on $\mathcal{T}$ is defined by:

---

21. The proof is the following:.
  - $x'_1$ contains only even integers.
  - For each k, $x'_k \neq 0^\infty$ since $x'_k$ is the result of the addition of a tuple and an integer.
  - If $x'_1$ is not empty, let $e_1$ denote the least even integer present in $x'_1$. As $x'_1 = x'_2 \oplus 2$, $x'_2$ is not empty and $e_2$ will denote the least integer present in $x'_2$. We have $e_1 = e_2 + 2$. So, we are able to build a sequence of positive even integers $e_1, e_2, \ldots$, which is strictly decreasing. That is impossible. So, $x'_1 = ()$.

**Definition 13 (Lexicographic ordering on tuples)**
Let $(x_1, \ldots, x_n, \ldots)$ and $(y_1, \ldots, y_m, \ldots)$ be 2 finite or infinite tuples $\in \mathcal{T}$.
$(x_1, \ldots, x_n, \ldots) <_{lex\infty} (y_1, \ldots, y_m, \ldots)$ iff $\exists i \geq 1$ such that:

- $\forall j < i$, $x_j = y_j$ and
- $y_i$ exists and:
    - either the tuple $(x_1, \ldots, x_n, \ldots)$ is finite with a number of elements equal to $i-1$ (so, $x_i$ does not exist),
    - or $x_i$ exists and $x_i < y_i$.

$(x_1, \ldots, x_n, \ldots) =_{lex\infty} (y_1, \ldots, y_m, \ldots)$ iff the tuples contain the same number $p \in \mathbb{N} \cup \{\infty\}$ of elements and $\forall i$, $1 \leq i \leq p$, $x_i = y_i$.
So, we define: $(x_1, \ldots, x_n, \ldots) \leq_{lex\infty} (y_1, \ldots, y_m, \ldots)$ iff
$\quad (x_1, \ldots, x_n, \ldots) =_{lex\infty} (y_1, \ldots, y_m, \ldots)$ or $(x_1, \ldots, x_n, \ldots) <_{lex\infty} (y_1, \ldots, y_m, \ldots)$.

The ordering $<_{lex\infty}$ is a generalisation of the classical lexicographic ordering (see Xuong, 1992) to the case of infinite tuples. This ordering is complete but not well-founded (there exist infinite sequences which are strictly non-increasing: $(0) <_{lex\infty} (0,0) <_{lex\infty} \ldots <_{lex\infty} (0, \ldots, 0, \ldots) <_{lex\infty} \ldots <_{lex\infty} (0,1)$).

Since the even values and the odd values in the tupled value of an argument do not play the same role, we cannot use a classical lexicographic comparison. So, we compare tupled values in two steps:

- The "first step" *compares the number of attack branches and the number of defence branches* of each argument. So, we have two criteria (one for the defence and the other for the attack). These criteria are aggregated using a *cautious method*: we conclude if one of the arguments has more defence branches (it is better according to the defence criterion) and less attack branches than the other argument (it is also better according to the attack criterion). Note that we conclude positively only when *all* the criteria agree: if one of the arguments has more defence branches (it is better according to the defence criterion) and more attack branches than the other argument (it is worse according to the attack criterion), the arguments are considered to be incomparable.
- Else, the arguments have the same number of defence branches and the same number of attack branches, and a "second step" *compares the quality of the attacks and the quality of the defences* using the length of each branch. This comparison is made with a lexicographic principle (see Definition 13) and gives two criteria which are again aggregated using a cautious method. In case of disagreement, the arguments are considered to be incomparable.

Let us consider some examples:

- $[(2), (1)]$ is better than $[(2), (1,1)]$ because there are less attack branches in the first tupled value than in the second tupled value, the numbers of defence branches being the same (first step).
- $[(2), (1)]$ is incomparable with $[(2,2), (1,1)]$ because there are less defence branches and less attack branches in the first tupled value than in the second tupled value (first step).

- $[(2),(3)]$ is better than $[(2),(1)]$ because there are weaker attack branches in the first tupled value than in the second tupled value (the attack branch of the first tupled value is longer than the one of the second tupled value), the defence branches being the same (second step, using the lexicographic comparison applied on even parts then on odd parts of the tupled values).
- $[(2),(3)]$ is better than $[(4),(3)]$ because there are stronger defence branches in the first tupled value than in the second tupled value (the defence branch is shorter in the first tupled value than in the second tupled value), the attack branches being the same (second step).
- $[(2),(1)]$ is incomparable with $[(4),(3)]$ because there are worse attack branches and better defence branches in the first tupled value than in the second tupled value (second step).

The comparison of arguments is done using Algorithm 1 which implements the principle of a double comparison (first quantitative, then qualitative) with two criteria (one defence criterion and one attack criterion) using a cautious method.

**Algorithm 1:** Comparison of two tupled values

% **Description of the parameters:** %
% $v$, $w$: 2 tupled values %
% **Notations:** %
%     $|v_p|$ (resp. $|w_p|$): number of elements in the even component of $v$ (resp. $w$) %
%         if $v_p$ (resp. $w_p$) is infinite then $|v_p|$ (resp. $|w_p|$) is taken equal to $\infty$ %
%     $|v_i|$ (resp. $|w_i|$): number of elements in the odd component of $v$ (resp. $w$) %
%         if $v_i$ (resp. $w_i$) is infinite then $|v_i|$ (resp. $|w_i|$) is taken equal to $\infty$ %
%     As usual, $\succ$ will denote the strict relation associated with $\succeq$ defined by: %
%         $v \succ w$ iff $v \succeq w$ and not$(w \succeq v)$. %

**begin**
1   **if** $v = w$ **then** $v \succeq w$ AND $w \succeq v$     % Case 1 %
2   **else**
3     **if** $|v_i| = |w_i|$ AND $|v_p| = |w_p|$ **then**
      % lexicographic comparisons between $v_p$ and $w_p$ and between $v_i$ and $w_i$ %
4       **if** $v_p \leq_{lex\infty} w_p$ AND $v_i \geq_{lex\infty} w_i$ **then** $v \succ w$     % case 2 %
5       **else**
6         **if** $v_p \geq_{lex\infty} w_p$ AND $v_i \leq_{lex\infty} w_i$ **then** $v \prec w$     % case 3 %
7         **else** $v \not\succeq w$ AND $v \not\preceq w$     % Incomparable tupled values. case 4 %
8     **else**
9       **if** $|v_i| \geq |w_i|$ AND $|v_p| \leq |w_p|$ **then** $v \prec w$     % case 5 %
10      **else**
11        **if** $|v_i| \leq |w_i|$ AND $|v_p| \geq |w_p|$ **then** $v \succ w$     % case 6 %
12        **else** $v \not\succeq w$ AND $v \not\preceq w$     % Incomparable tupled values. Case 7 %
**end**

Algorithm 1 defines a partial preordering on the set $v(\mathcal{A})$:

**Property 9 (Partial preordering)** *Algorithm 1 defines a partial preordering $\succeq$ on the set $v(\mathcal{A})$.*

*The tupled value $[0^\infty, ()]$ is the only maximal value of the partial preordering $\succeq$.*
*The tupled value $[(), 1^\infty]$ is the only minimal value of the partial preordering $\succeq$.*

**Notation:** the partial preordering $\succeq$ on the set $v(\mathcal{A})$ induces a partial preordering on the arguments (the partial preordering on $\mathcal{A}$ will be denoted like the partial preordering on $v(\mathcal{A})$): $A \succeq B$ if and only if $v(A) \succeq v(B)$[22].

In order to present the underlying principles satisfied by the global valuation, we first consider the different ways for modifying the defence part or the attack part of an argument:

**Definition 14 (Adding/removing a branch to an argument)**
*Let $A$ be an argument whose tupled value is $v(A) = [v_p(A), v_i(A)]$ with $v_p(A) = (x_1^p, \ldots, x_n^p)$ and $v_i(A) = (x_1^i, \ldots, x_m^i)$ ($v_p(A)$ or $v_i(A)$ may be empty but not simultaneously).*
*Adding (resp. removing) a defence branch to $A$ is defined by:*
*$v_p(A)$ becomes $\text{Sort}(x_1^p, \ldots, x_n^p, x_{n+1}^p)$ where $x_{n+1}^p$ is the length of the added branch (resp. $\exists j \in [1..n]$ such that $v_p(A)$ becomes $(x_1^p, \ldots, x_{j-1}^p, x_{j+1}^p, \ldots, x_n^p)$).*
*And the same thing on $v_i(A)$ for adding (resp. removing) an attack branch to $A$.*

**Definition 15 (Increasing/decreasing the length of a branch of an argument)**
*Let $A$ be an argument whose tupled value is $v(A) = [v_p(A), v_i(A)]$ with $v_p(A) = (x_1^p, \ldots, x_n^p)$ and $v_i(A) = (x_1^i, \ldots, x_m^i)$ ($v_p(A)$ or $v_i(A)$ may be empty but not simultaneously).*
*Increasing (resp. decreasing) the length of a defence branch of $A$ is defined by:*
*$\exists j \in [1..n]$ such that $v_p(A)$ becomes $(x_1^p, \ldots, x_{j-1}^p, x_j'^p, x_{j+1}^p, \ldots, x_n^p)$ where $x_j'^p > x_j^p$ (resp. $x_j'^p < x_j^p$) and the parity of $x_j'^p$ is the parity of $x_j^p$.*
*And the same thing on $v_i(A)$ for increasing (resp. decreasing) an attack branch to $A$.*

**Definition 16 (Improvement/degradation of the defences/attacks)**
*Let $A$ be an argument whose tupled value is $v(A) = [v_p(A), v_i(A)]$ ($v_p(A)$ or $v_i(A)$ may be empty but not simultaneously). We define:*

**An improvement (resp. degradation) of the defence** *consists in*
- *adding a defence branch to $A$ if initially $v_p(A) \neq 0^\infty$ (resp. removing a defence branch of $A$);*
- *or decreasing (resp. increasing) the length of a defence branch of $A$;*
- *or removing the only defence branch leading to $A$ (resp. adding a defence branch leading to $A$ if initially $v_p(A) = 0^\infty$);*

**An improvement (resp. degradation) of the attack** *consists in*
- *adding (resp. removing) an attack branch to $A$;*
- *or decreasing (resp. increasing) the length of an attack branch of $A$.*

**Property 10 (Underlying principles)** *Let $v$ be a valuation with tuples (Definition 10) associated with Algorithm 1, $v$ respects the following principles:*

**P1'** *The valuation is maximal for an argument without attackers and non maximal for an argument which is attacked (whether it is defended or not).*

---

22. We will also use the notation $B \preceq A$ defined by: $B \preceq A$ iff $A \succeq B$.

**P2′** *The valuation of an argument takes into account all the branches which are rooted in this argument.*

**P3′** *The improvement of the defence or the degradation of the attack of an argument leads to an increase of the value of this argument.*

**P4′** *The improvement of the attack or the degradation of the defence of an argument leads to a decrease of the value of the argument.*

**Example 4 (continuation)** *With the valuation with tuples, we obtain:*

- $v(E_1) = v(D_2) = v(D_3) = v(C_4) = v(B_4) = [0^\infty, ()]$,
- $v(D_1) = v(C_2) = v(C_3) = v(B_3) = [(), (1)]$,
- $v(C_1) = v(B_2) = [(2), ()]$,
- $v(B_1) = [(2), (3)]$,
- $v(A) = [(2, 4), (1, 3, 3)]$.

*So, we have:*

$$\begin{array}{cc}
\begin{array}{c} E_1, D_2, D_3, C_4, B_4 \\ \succ \\ C_1, B_2 \\ \succ \\ B_1 \\ \succ \\ D_1, C_2, C_3, B_3 \end{array} & \text{but also} \quad \begin{array}{c} E_1, D_2, D_3, C_4, B_4 \\ \succ \\ A \end{array}
\end{array}$$

*A is incomparable with almost all the other arguments (except with the leaves of the graph). Similarly, on the hatched part of the graph, we obtain the following results:*

$$E_1, D_2 \succ C_1 \succ B_1 \succ A \succ D_1, C_2$$

*A is now comparable with all the other arguments (in particular, A is "worse" than its defender $C_1$ and than its direct attacker $B_1$).*

### 3.3 Main differences between "local" and "global" valuations

Cayrol and Lagasquie-Schiex (2003c) give a comparison of these approaches with some existing approaches (Dung, 1995; Jakobovits & Vermeir, 1999; Besnard & Hunter, 2001), and also a comparison of the "local" approaches and the "global" approach. The improvement of the global approach proposed in this paper does not modify the main results of this comparison.

Let us recall here an example of the essential point which differentiates them (this example has already been presented at the beginning of Section 3):

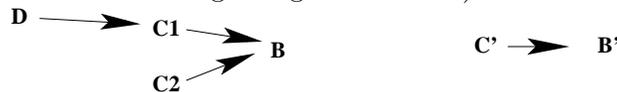

In the local approach, $B'$ is better than $B$ (since $B'$ suffers one attack whereas $B$ suffers two attacks).

In the global approach, $B$ is better than $B'$ (since it has at least a defence whereas $B'$ has none). In this case, $C_1$ loses its negative status of attacker, since it is in fact "carrying a defence" for $B$.

The following table synthesises the results about the different proposed valuations:

*global approach*

| arguments having only attack branches | $\preceq$ | arguments having attack branches and defence branches | $\preceq$ | arguments having only defence branches | $\preceq$ | arguments never attacked |
|---|---|---|---|---|---|---|

*local approach*

| arguments having several unattacked direct attackers | $\preceq$ | arguments having only one unattacked direct attacker | $\preceq$ | arguments having only one attacked direct attacker (possibly defended) | $\preceq$ | arguments never attacked |
|---|---|---|---|---|---|---|
| arguments having several attacked direct attackers (possibly defended) | | | | | | |

The difference between the local approaches and the global approach is also illustrated by the following property:

**Property 11 (Independence of branches in the global approach)**
*Let $A$ be an argument having the following direct attackers:*

- $A_1$ whose value is $v(A_1) = [(a^1_{p_1}, \ldots, a^1_{p_{m_1}}), (a^1_{i_1}, \ldots, a^1_{i_{m_1}})]$,
- $\ldots$,
- $A_n$ whose value is $v(A_n) = [(a^n_{p_1}, \ldots, a^n_{p_{m_n}}), (a^n_{i_1}, \ldots, a^n_{i_{m_n}})]$.

*Let $A'$ be an argument having the following direct attackers:*

- $A^1_{p_1}$ whose value is $v(A^1_{p_1}) = [(a^1_{p_1})()]$,
- $\ldots$,
- $A^1_{p_{m_1}}$ whose value is $v(A^1_{p_{m_1}}) = [(a^1_{p_{m_1}})()]$,
- $A^1_{i_1}$ whose value is $v(A^1_{i_1}) = [()(a^1_{i_1})]$,
- $\ldots$,
- $A^1_{i_{m_1}}$ whose value is $v(A^1_{i_{m_1}}) = [()(a^1_{i_{m_1}})]$,
- $\ldots$,
- $A^n_{p_1}$ whose value is $v(A^n_{p_1}) = [(a^n_{p_1})()]$,
- $\ldots$,

- $A_{p_{m_n}}^n$ whose value is $v(A_{p_{m_n}}^n) = [(a_{p_{m_n}}^n)()]$,
- $A_{i_1}^n$ whose value is $v(A_{i_1}^n) = [()(a_{i_1}^n)]$,
- ...,
- $A_{i_{m_n}}^n$ whose value is $v(A_{i_{m_n}}^n) = [()(a_{i_{m_n}}^n)]$.

Then $v(A) = v(A')$.

This property illustrates the "independence" of branches during the computation of the values in the global approach, even when these branches are not graphically independent. On the following example, $A$ and $A'$ have the same value $[(2,2)()]$ though they are the root of different subgraphs:

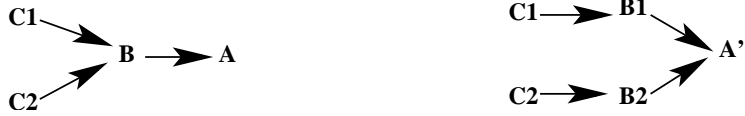

This property is not satisfied by the local approach since, using the underlying principles of the local approach (see Property 4), the value of the argument $A$ must be at least as good as (and sometimes better than[23]) the value of the argument $A'$ ($A$ having one direct attacker, and $A'$ having two direct attackers).

### 3.4 Conclusion about valuation step

We have proposed two different gradual valuation models and we are now able to make a distinction between different arguments using the preordering associated with a valuation model. These valuations will be used for the selection of the arguments (see Section 4).

## 4. Graduality and acceptability

In this section, we now shift to the selection step and introduce graduality in the notion of acceptability[24].

The basic idea is to select an argument depending on the non-selection of its direct attackers. Following this idea, we propose two different methods:

- The first method consists in refining the classical partition issued from Dung's collective acceptability; this refinement may be achieved using the gradual valuations defined in Section 3.
- The second method takes place in an individual acceptability and consists in defining a new acceptability using only the gradual valuations defined in Section 3.

### 4.1 Dung's (1995) collective acceptability

In the framework of collective acceptability, we have to consider the acceptability of a set of arguments. This acceptability is defined with respect to some properties and the sets which satisfy these properties are called *acceptable sets* or *extensions*. An argument will be said acceptable if and only if it belongs to an extension.

---

23. With the valuation proposed by Besnard and Hunter (2001), we obtain: $v(A) = \frac{3}{4}$ and $v(A') = \frac{1}{2}$.
24. This work has been presented in a workshop (Cayrol & Lagasquie-Schiex, 2003b).

**Definition 17 (Basic properties of extensions following Dung, 1995)**
Let $<\mathcal{A}, \mathcal{R}>$ be an argumentation system, we have:

**Conflict-free set** A set $E \subseteq \mathcal{A}$ is conflict-free if and only if $\nexists A, B \in E$ such that $A\mathcal{R}B$.

**Collective defence** Consider $E \subseteq \mathcal{A}$, $A \in \mathcal{A}$. $E$ collectively defends $A$ if and only if $\forall B \in \mathcal{A}$, if $B\mathcal{R}A$, $\exists C \in E$ such that $C\mathcal{R}B$. $E$ defends all its elements if and only if $\forall A \in E$, $E$ collectively defends $A$.

Dung (1995) defines several semantics for collective acceptability: mainly, the *admissible semantics*, the *preferred semantics* and the *stable semantics* (with corresponding extensions: the admissible sets, the preferred extensions and the stable extensions).

**Definition 18 (Some semantics and extensions following Dung, 1995)** Let $<\mathcal{A}, \mathcal{R}>$ be an argumentation system.

**Admissible semantics (admissible set)** A set $E \subseteq \mathcal{A}$ is admissible if and only if $E$ is conflict-free and $E$ defends all its elements.

**Preferred semantics (preferred extension)** A set $E \subseteq \mathcal{A}$ is a preferred extension if and only if $E$ is maximal for set inclusion among the admissible sets.

**Stable semantics (stable extension)** A set $E \subseteq \mathcal{A}$ is a stable extension if and only if $E$ is conflict-free and $E$ attacks each argument which does not belong to $E$ ($\forall A \in \mathcal{A} \setminus E$, $\exists B \in E$ such that $B\mathcal{R}A$).

Note that in all the above definitions, *each attacker* of a given argument is considered separately (the "direct attack" as a whole is not considered). Dung (1995) proves that:

- Any admissible set of $<\mathcal{A}, \mathcal{R}>$ is included in a preferred extension of $<\mathcal{A}, \mathcal{R}>$.
- There always exists at least one preferred extension of $<\mathcal{A}, \mathcal{R}>$.
- If $<\mathcal{A}, \mathcal{R}>$ is well-founded then there is only one preferred extension which is also the only stable extension.
- Any stable extension is also a preferred extension (the converse is false).
- There is not always a stable extension.

**Property 12** *The set of leaves (i.e. $\{A | \mathcal{R}^-(A) = \varnothing\}$) is included in every preferred extension and in every stable extension.*

### 4.2 Different levels of collective acceptability

Under a given semantics, and following Dung, the acceptability of an argument depends on its membership to an extension under this semantics. We consider three possible cases[25]:

---

25. The terminology used in this section is also used in the domain of nonmonotonic reasoning (see Pinkas & Loui, 1992): the word *uni* comes from the word *universal* which is a "synonym" of the word *skeptical*, and the word *exi* comes from the word *existential* which is a "synonym" of the word *credulous*. We have chosen to use the words *uni* and *exi* because they recall the logical quantificators $\forall$ (*for all*) and $\exists$ (*exists at least one*).

- the argument can be *uni-accepted*, when it belongs to all the extensions of this semantics,
- or the argument can be *exi-accepted*, when it belongs to at least one extension of this semantics,
- or the argument can be *not-accepted* when it does not belong to any extension of this semantics.

However, these three levels seem insufficient. For example, what should be concluded in the case of two arguments $A$ and $B$ which are exi-accepted and such that $A\mathcal{R}B$ or $B\mathcal{R}A$?

So, we introduce a new definition which takes into account the situation of the argument w.r.t. its attackers. This refines the class of the exi-accepted arguments under a given semantics $S$.

**Definition 19 (Cleanly-accepted argument)** *Consider $A \in \mathcal{A}$, $A$ is* cleanly-accepted *if and only if $A$ belongs to at least one extension of $S$ and $\forall B \in \mathcal{A}$ such that $B\mathcal{R}A$, $B$ does not belong to any extension of $S$.*

Thus, we capture the idea that an argument will be better accepted, if its attackers are not-accepted.

**Property 13** *Consider $A \in \mathcal{A}$ and a semantics $S$ such that each extension for $S$ is conflict-free. If $A$ is uni-accepted then $A$ is cleanly-accepted. The converse is false.*

The notion of cleanly-accepted argument refines the class of the exi-accepted arguments. For a semantics $S$ and an argument $A$, we have the following states:

- $A$ can be *uni-accepted*, if $A$ belongs to all the extensions for $S$ (so, it will also be cleanly-accepted);
- or $A$ can be *cleanly-accepted* (so, it is by definition also exi-accepted); note that it is possible that the argument is also uni-accepted;
- or $A$ can be *only-exi-accepted*, if $A$ is not cleanly-accepted, but $A$ is exi-accepted;
- or $A$ is *not-accepted* if $A$ does not belong to any extension for $S$.

**Example 9** *Consider the following argumentation system.*

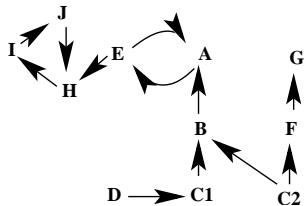

There are two preferred extensions $\{D, C_2, A, G\}$ and $\{D, C_2, E, G, I\}$. So, for the preferred semantics, the acceptability levels are the following:

- $D$, $C_2$ and $G$ are uni-accepted,
- $I$ is cleanly-accepted but not uni-accepted,
- $A$ and $E$ are only-exi-accepted,
- $B$, $C_1$, $F$, $H$ and $J$ are not-accepted.

Note that, in all the cases where there is only one extension, the first three levels of acceptability coincide[26]. This is the case:

---

26. If there is only one extension then the fact that $A$ belongs to all the extensions is equivalent to the fact that $A$ belongs to at least one extension. Moreover, with only one extension containing $A$, all the attackers of $A$ do not belong to an extension. So, $A$ is cleanly-accepted.

- Under the preferred semantics, when there is no even cycle (see Doutre, 2002).
- Under the basic semantics (another semantics proposed by Dung – see Dung, 1995; Doutre, 2002 – which is not presented here and which has only one extension).

Looking more closely, we can prove the following result (proof in Appendix A):

**Property 14** *Under the stable semantics, the class of the uni-accepted arguments coincides with the class of the cleanly-accepted arguments.*

Then, using a result issued from the work of Dunne and Bench-Capon (2001, 2002) and reused by Doutre (2002) which shows that, when there is no odd cycle, all the preferred extensions are stable[27], we apply Property 14 and we obtain the following consequence:

**Consequence 1** *Under the preferred semantics, when there is no odd cycle, the class of the uni-accepted arguments coincides with the class of the cleanly-accepted arguments.*

Finally, the exploitation of the gradual interaction-based valuations (see Section 3) allows us to define new levels of collective acceptability.

Let $v$ be a gradual valuation and let $\succeq$ be the associated preordering (partial or complete) on $\mathcal{A}$. This preordering can be used inside each acceptability level (for example, the level of the exi-accepted arguments) in order to identify arguments which are better accepted than others.

**Example 9 (continuation)** *Two different gradual valuations are applied on the same graph:*

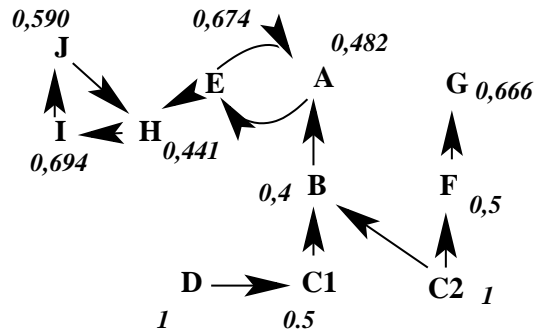

**Besnard & Hunter's (2001) valuation**

*With the instance of the generic valuation proposed by Besnard and Hunter (2001) (see Section 3.1), we obtain the following comparisons:*

$$D, C_2 \succ I \succ E \succ G \succ J \succ C_1, F \succ A \succ H \succ B$$

---
27. This corresponds to the consistent argumentation system proposed by Dung (1995).

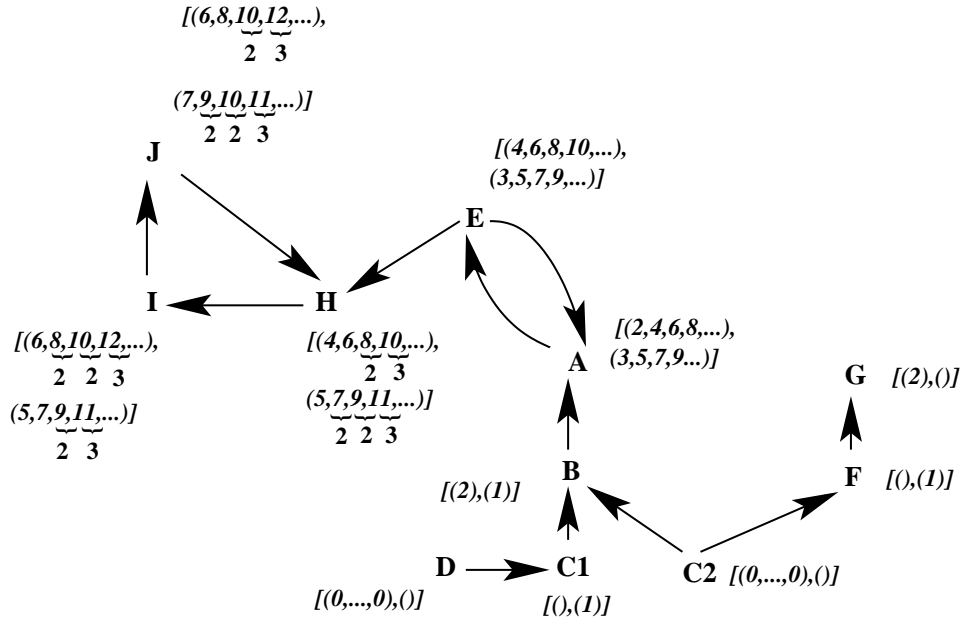

**Valuation with tuples**

*With the global valuation with tuples presented in Section 3.2, we obtain the following comparisons:*

$$D, C_2 \succ G \succ B \succ F, C_1$$
$$D, C_2 \succ A \succ E$$
$$D, C_2 \succ H \succ E$$
$$D, C_2 \succ I$$
$$D, C_2 \succ J$$

*So, all the arguments belonging to a cycle are incomparable with $G$, $B$, $F$, $C_1$ and, even between them, there are few comparison results.*

If we apply the preordering induced by a valuation without respecting the acceptability levels defined in this section, counter-intuitive situations may happen. In Example 9, we obtain:

- With the valuation of Besnard and Hunter (2001) and under the preferred semantics, $E \succ G$ despite the fact that $G$ is uni-accepted and $E$ is only-exi-accepted.
- With the valuation with tuples and under the preferred semantics, $H \succ E$ despite the fact that $E$ is only-exi-accepted and $H$ is not-accepted.

These counter-intuitive situations illustrate the difference between the acceptability definition and the valuation definitions (even if both use the interaction between arguments, they do not use it in the same way).

275

### 4.3 Towards a gradual individual acceptability

The individual acceptability is based on the comparison of an argument with its attackers. The first proposal has been to select an argument if and only if it does not have any attacker (see Elvang-Goransson et al., 1993).

This has later been extended by Amgoud and Cayrol (1998) where, using a preference relation between arguments (an intrinsic valuation), an argument is accepted if and only if it is preferred to each of its attackers.

Following this proposal, we propose the same mechanism but with the *interaction-based valuation*.

Given $v$ a gradual valuation, the preordering induced by $v$ can be directly used in order to compare, from the acceptability point of view, an argument and its attackers[28]. This defines a new class of acceptable arguments: well-defended arguments.

**Definition 20 (Well-defended argument)** *Consider $A \in \mathcal{A}$, $A$ is* well-defended *(for $v$) if and only if $\forall B \in \mathcal{A}$ such that $B\mathcal{R}A$, $B \not\succ A$.*

Thus, we capture the idea that an argument will be better accepted if it is at least as good as its direct attackers (or incomparable with them in the case of a partial ordering). The set of well-defended arguments will depend on the valuation used.

Using this new notion, the set of the arguments is partitioned in three classes:

- the first class contains the arguments which are not attacked,
- the second class contains the arguments which are attacked but are well-defended,
- the third class contains the other arguments (attacked and not well-defended).

Note that the set of the well-defended arguments corresponds to the union of the two first classes. A further refinement uses the gradual valuation inside each of the classes as in Section 4.2.

In Example 9 presented in Section 4.2, the well-defended arguments are:

- $D$, $C_2$, $G$, $H$ and $A$ ($A$ is incomparable with $B$ but better than $E$) for the valuation with tuples,
- though with the valuation of Besnard and Hunter (2001) the well-defended arguments are $D$, $C_2$, $G$, $I$ and $E$ ($E$ is better than $A$).

Note also that, as in the semantics of Dung (1995), Definition 20 considers the attackers one by one. It is not suitable for a valuation which handles the "direct attack" as a whole (as the valuation of Besnard and Hunter (2001) – see the counterexamples presented in Section 4.4).

---

28. This idea is also used in the notion of "defeat" proposed by Bench-Capon (2002). So, there is a link between a "well-defended argument" and an argument which is not "attacked" in the sense of Bench-Capon (2002) by its direct attackers. Note that, in the work of Bench-Capon (2002), the valuation is an extra knowledge added in the argumentation framework. In contrast, here, the $v$-preference is extracted from the attack graph.

### 4.4 Compatibility between acceptability and gradual valuation

Following the previous sections, the set of arguments can be partitioned in two different ways:

- First, given a semantics $S$ and a gradual valuation $v$, it is possible to use the partition issued from Dung (1995) which we have refined:

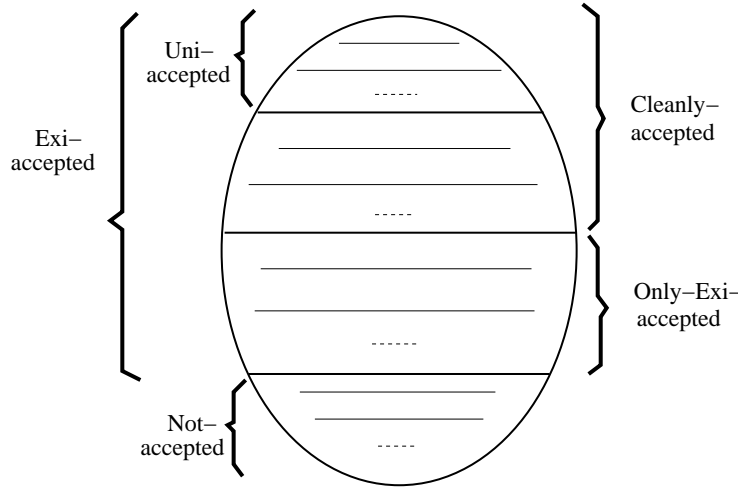

*Refinement of each level with the gradual valuation v*

- Second, given a gradual valuation $v$, it is possible to use the partition induced by the notion of well-defended arguments:

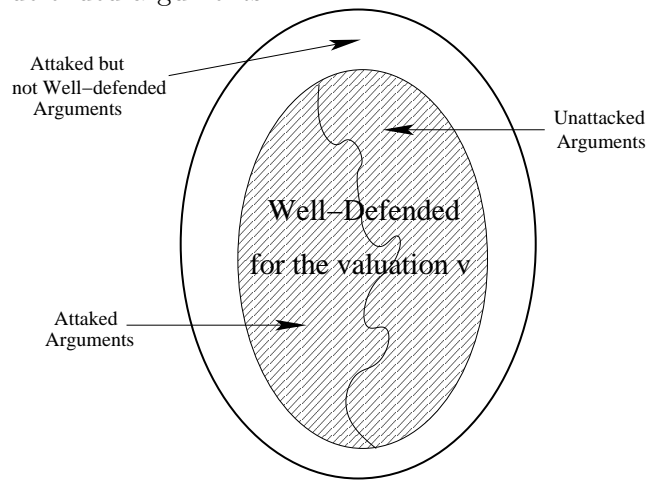

A very natural and interesting question is: is it possible to find a semantics $S$ and a gradual valuation $v$ such that the associated partitions have some compatibilities?

The following examples show that the class of the well-defended arguments does not correspond to the class of the cleanly-accepted arguments (in some cases, some uni-accepted arguments are even not well-defended).

4.4.1 Examples showing the non-compatibility in the general case

We give examples for each usual valuation (the global valuation with tuples and 2 instances of the generic local valuation: Besnard & Hunter, 2001; Jakobovits & Vermeir, 1999) and for the most classical semantics for acceptability (preferred semantics and stable semantics of Dung, 1995).

*Cleanly-accepted argument but not well-defended:* There are 3 examples (each using a distinct valuation: one for the global valuation and two for the two well-known instances of the local valuation):

- the argument $A$ is cleanly-accepted but it is not well-defended:

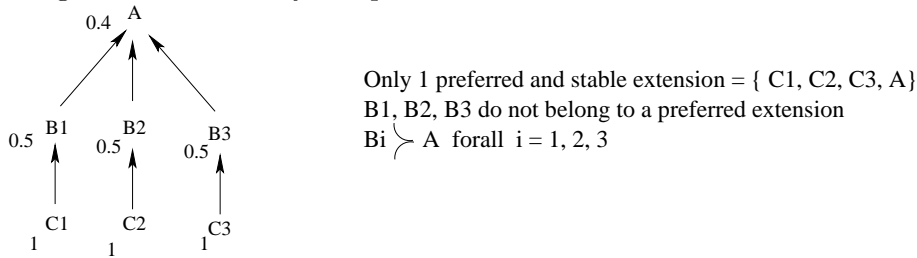

Only 1 preferred and stable extension = { C1, C2, C3, A}
B1, B2, B3 do not belong to a preferred extension
$B_i \succ A$ forall $i = 1, 2, 3$

- the argument $A$ is cleanly-accepted but it is not well-defended:

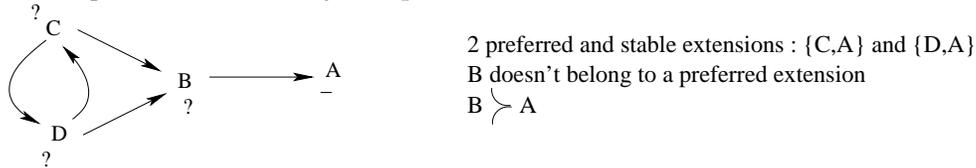

2 preferred and stable extensions : {C,A} and {D,A}
B doesn't belong to a preferred extension
$B \succ A$

- the argument $I$ is cleanly-accepted but it is not well-defended:

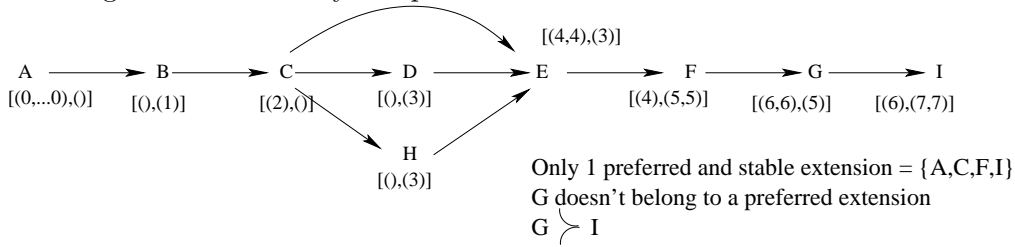

Only 1 preferred and stable extension = {A,C,F,I}
G doesn't belong to a preferred extension
$G \succ I$

*Well-defended argument but not cleanly-accepted:* Similarly, for the same three valuations, we have:

- the argument $C$ is well-defended but it is not cleanly-accepted:

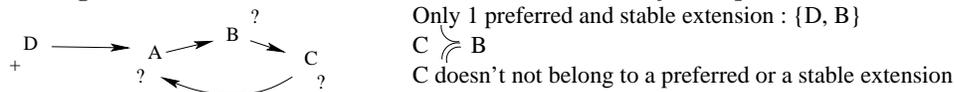

Only 1 preferred and stable extension : {D, B}
$C \succ B$
C doesn't not belong to a preferred or a stable extension

- the argument $F$ is well-defended but it is not cleanly-accepted:

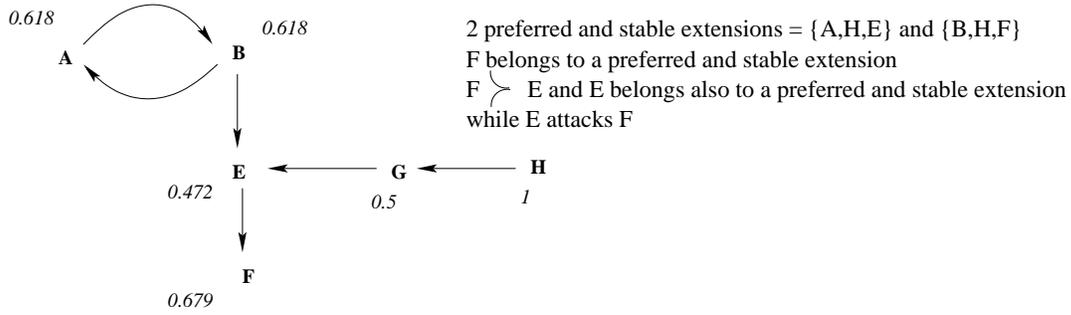

- the argument $G$ is well-defended but it is not cleanly-accepted:

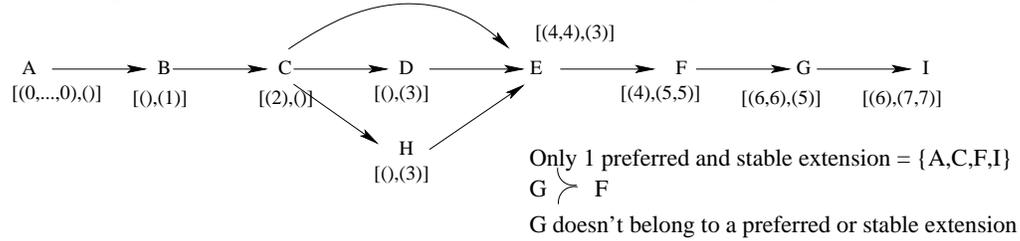

Only 1 preferred and stable extension = {A,C,F,I}
$G \succ F$
G doesn't belong to a preferred or stable extension

### 4.4.2 Particular cases leading to compatibility

In the context of an argumentation system with a finite relation $\mathcal{R}$ without cycles[29], the stable and the preferred semantics provide only one extension and the levels of uni-accepted, exi-accepted, cleanly-accepted coincide.
In this context, there are at least two particular cases leading to compatibility.

*First case:* It deals with the global valuation with tuples.

**Theorem 1** *Let $\mathcal{G}$ be the graph associated with $<\mathcal{A},\mathcal{R}>$, $<\mathcal{A},\mathcal{R}>$ being an argumentation system with a finite relation $\mathcal{R}$ without cycles and satisfying the following condition: $\exists$ $A \in \mathcal{A}$ such that*

- $\forall X_i$, *leaf of $\mathcal{G}$, $\exists$ only one path from $X_i$ to $A$, $X_i^1 - \ldots - X_i^{l_i} - A$ with $X_i^1 = X_i$ and $l_i$ the length of this path (if $l_i$ is even, this path is a defence branch for $A$, else it is an attack branch)*,
- *all the paths from $X_i$ to $A$ are root-dependent in $A$*,
- $\forall A_i \in \mathcal{A}$, $\exists X_j$ *a leaf of $\mathcal{G}$ such that $A_i$ belongs to a path from $X_j$ to $A$.*

*Let $v$ be a valuation with tuples. Let $S$ be a semantics $\in$ {preferred, stable}.*

1. $\forall B \in \mathcal{A}$, $B \neq A$, $B$ *(exi, uni, cleanly) accepted for $S$ iff $B$ well-defended for $v$.*

2. *If $A$ is (exi, uni, cleanly) accepted for $S$ then $A$ is well-defended for $v$ (the converse is false).*

3. *If $A$ is well-defended for $v$ and if all the branches leading to $A$ are defence branches for $A$ then $A$ is (exi, uni, cleanly) accepted for $S$.*

---

29. So, $(\mathcal{A},\mathcal{R})$ is well-founded.

Note that Theorem 1 is, in general, not satisfied by a local valuation. See the following counterexample for the valuation of Besnard and Hunter (2001):

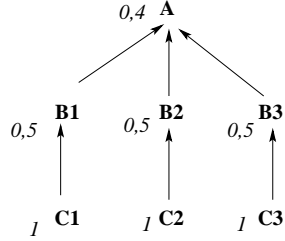

The graph satisfies the condition stated in Theorem 1. The set of well-defended arguments is $\{C_1, C_2, C_3\}$ (so, $A$ is not well-defended). Nevertheless, $\{C_1, C_2, C_3, A\}$ is the preferred extension.

*Second case:* This second case concerns the generic local valuation:

**Theorem 2** *Let $<\mathcal{A}, \mathcal{R}>$ be an argumentation system with a finite relation $\mathcal{R}$ without cycles. Let $S$ be a semantics $\in \{preferred, stable\}$. Let $v$ be a generic local valuation satisfying the following condition $(*)$:*
$$(\forall i = 1 \ldots n, g(x_i) \geq x_i) \Rightarrow (g(h(x_1, \ldots, x_n)) \geq h(x_1, \ldots, x_n)) \qquad (*)$$
*$\forall A \in \mathcal{A}$, $A$ (exi, uni, cleanly) accepted for $S$ iff $A$ well-defended for $v$.*

This theorem is a direct consequence of the following lemma:

**Lemma 1** *Let $<\mathcal{A}, \mathcal{R}>$ be an argumentation system with a finite relation $\mathcal{R}$ without cycles. Let $S$ be a semantics $\in \{preferred, stable\}$. Let $v$ be a generic local valuation satisfying the condition $(*)$.*

*(i) If $A$ is exi-accepted and $A$ has only one direct attacker $B$ then $A \succeq B$.*

*(ii) If $B$ is not-accepted and $B$ has only one direct attacker $C$ then $C \succeq B$.*

**Remark:** The condition $(*)$ stated in Theorem 2 is:

- false for the local valuation proposed by Besnard and Hunter (2001) as shown in the following graph:

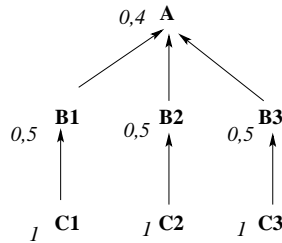

We know that $g(x) = \frac{1}{1+x}$ and $h(x_1, \ldots, x_n) = \Sigma_{i=1}^n x_i$ (see Property 6). We get:
- $\forall i = 1 \ldots 3$, $x_i = v(B_i) = 0.5$,
- $\forall i = 1 \ldots 3$, $g(x_i) = 0.66$, so $g(x_i) \geq x_i$,
- and nevertheless $g(h(x_1, x_2, x_3)) = v(A) = 0.4 \not\geq h(x_1, x_2, x_3) = 1.5$.

- false for the local valuations defined with $h$ such that $\exists n > 1$ with $h(x_1, \ldots, x_n) > \max(x_1, \ldots, x_n)$ (for all the functions $g$ strictly non-increasing): see the previous graph where $h(x_1, x_2, x_3) = 1.5$ and $\max(x_1, x_2, x_3) = 0.5$.
- true for the local valuations defined with $h = \max$ (for all the functions $g$): if $h = \max$ then $g(h(x_1, \ldots, x_n)) = g(\max(x_1, \ldots, x_n)) = g(x_j)$, $x_j$ being the maximum of the $x_i$; and, by assumption, $g(x_i) \geq x_i$, $\forall x_i$, so in particular for $x_j$; so, we get:
$$g(h(x_1, \ldots, x_n)) = g(x_j) \geq x_j = \max(x_1, \ldots, x_n) = h(x_1, \ldots, x_n).$$

## 5. Conclusion

In this paper, we have introduced graduality in the two main related issues of argumentation systems:

- the valuation of the arguments,
- the acceptability of the arguments.

Regarding the first issue, we have defined two formalisms introducing an interaction-based gradual valuation of arguments.

- First, a generic gradual valuation which covers existing proposals (for example Besnard & Hunter, 2001 and Jakobovits & Vermeir, 1999). This approach is essentially "local" since it computes the value of the argument only from the value of its direct attackers.
- Then, an approach based on a labelling which takes the form of a pair of tuples; this labelling memorises the structure of the graph representing the interactions (the "attack graph"), associating each branch with its length (number of the edges from the leaf to the current node) in the attack graph (if the length of the branch is an even integer, the branch is a defence branch for the current node, otherwise the branch is an attack branch for the current node). This approach is said to be "global" since it computes the value of the argument using the whole attack graph influencing the argument.

We have shown that each of these valuations induces a preordering on the set of the arguments, and we have brought to light the main differences between these two approaches.

Regarding the second issue, two distinct approaches have been proposed:

- First, in the context of the collective acceptability of Dung (1995): three levels of acceptability (uni-accepted, exi-accepted, not-accepted) were already defined. More graduality can be introduced in the collective acceptability using the notion of *cleanly-accepted* arguments (those whose direct attackers are not-accepted).
- Then, in the context of individual acceptability: using the previously defined gradual valuations, the new notion of *well-defended* arguments has been introduced (those which are preferred to their direct attackers in the sense of a given gradual valuation $v$).

The first concept induces a refinement of the level of exi-accepted in two sublevels (cleanly-accepted arguments and only-exi-accepted arguments). The gradual valuation allows graduality inside each level of this collective acceptability.

The second concept induces two new levels of acceptability (well-defended arguments and not-well-defended arguments). The gradual valuation also allows graduality inside each level of this individual acceptability.

Regarding our initial purpose of introducing graduality in the definition of acceptability, we have adopted a basic principle:

- acceptability is strongly related to the interactions between arguments (represented on the graph of interactions),
- and an argument is all the more acceptable if it is preferred to its direct attackers.

Then, we have followed two different directions. One is based on a refinement of an existing partition and remains in the framework of Dung's work. The other one is based on the original concept of "being well-defended", and deserves further investigation, in particular from a computational point of view.

## Acknowledgements

Thanks to the reviewers for their very interesting and constructive comments.
Thanks to Thomas SCHIEX for his help.

## Appendix A. The proofs

In this section, we give the proofs of all the properties presented in Sections 3 and 4.

> **Proof**
> **(of Property 1)** By induction from $V_{\text{Min}} \leq g(V_{\text{Max}}) < V_{\text{Max}}$ and by applying function $g$ twice. □
>
> **Proof**
> **(of Property 2)** The valuation function $v$ associates each argument $A$ with a value $v(A)$ belonging to a set $V$ which is a subset of a completely ordered set $W$. □
>
> **Proof**
> **(of Property 3)** Let $\mathcal{C} = A_n - A_{n-1} - \ldots - A_2 - A_1$ be a cycle:
>
> - If $n$ is even: $n = 2k$ and $v(A_1) = g(v(A_2)) = \ldots = g^{2k-1}(v(A_{2k})) = g^{2k}(v(A_1))$; so, $v(A_1)$ is a fixpoint of $g^{2k} = g^n$. It is the same for each $A_i$, $1 \leq i \leq 2k$.
>   However, the $A_i$ may have different values: for example, for $n = 2$, with the valuation of Jakobovits and Vermeir (1999), $v(A_1) = +$ and $v(A_2) = -$ with $g(+) = -$ and $g(-) = +$. If all the $A_i$ have the same value, then this value will be a fixpoint of $g$ (because $v(A_1) = g(v(A_2)) = g(v(A_1))$).

- If $n$ is odd: $n = 2k + 1$ and $v(A_1) = g(v(A_2)) = \ldots = g^{2k}(v(A_{2k+1})) = g^{2k+1}(v(A_1))$; so, $v(A_1)$ is a fixpoint of $g^{2k+1} = g^n$. It is the same for each $A_i$, $1 \leq i \leq 2k + 1$.

  Since the function $g$ is non-increasing, the function $g^{2k+1}$ is also non-increasing and we can apply the following result: "if a non-increasing function has fixpoints, these fixpoints are identical"[30]. So, $v(A_1) = \ldots = v(A_{2k+1})$. But, $v(A_1) = g(v(A_2)) = g(v(A_1))$, so $v(A_1)$ is a fixpoint of $g$.
  So, for all the $1 \leq i \leq 2k + 1$, $v(A_i)$ is a fixpoint of $g$.

  $\square$

**Proof**
**(of Property 4)**
**P1** is satisfied because: $\forall A \in \mathcal{A}$, if $A$ has no direct attacker ($\mathcal{R}^-(A)$ is empty), then $v(A) = V_{\text{Max}}$ and $g(V_{\text{Max}}) < V_{\text{Max}}$.

**P2** is satisfied because if $\mathcal{R}^-(A) = \{A_1, \ldots, A_n\}$, $h(v(A_1), \ldots, v(A_n))$ evaluates the "direct attack" of $A$.

**P3** is satisfied because the function $g$ is supposed to be non-increasing.

**P4** is satisfied due to the properties of the function $h$.   $\square$

**Proof**
**(of Property 5)** The valuation proposed by Jakobovits and Vermeir (1999) is the following:

Let $<\mathcal{A}, \mathcal{R}>$ be an argumentation system. A complete labelling of $<\mathcal{A}, \mathcal{R}>$ is a function $Et : \mathcal{A} \to \{+, ?, -\}$ such that:

1. If $Et(A) \in \{?, -\}$ then $\exists B \in \mathcal{R}^-(A)$ such that $Et(B) \in \{+, ?\}$
2. If $Et(A) \in \{+, ?\}$ then $\forall B \in \mathcal{R}^-(A)$ or $\in \mathcal{R}^+(A)$, $Et(B) \in \{?, -\}$

Moreover, Jakobovits and Vermeir (1999) also define a complete rooted labelling $Et$ with: $\forall A \in \mathcal{A}$, if $Et(A) = -$ then $\exists B \in \mathcal{R}^-(A)$ such that $Et(B) = +$.

The translation of $Et$ into a local gradual valuation is very easy:

$g$ is defined by $g(-) = +$, $g(+) = -$, $g(?) = ?$ and $h$ is the function max.   $\square$

**Proof**
**(of Property 6)** Besnard and Hunter (2001) introduce the following function $Cat$ (in the context of "deductive" arguments and for an acyclic graph):

- if $\mathcal{R}^-(A) = \emptyset$, then $Cat(A) = 1$

---

30. Proof: let $g$ be a non-increasing function, let $\alpha$ and $\beta$ be two fixpoints of $g$. If $\alpha \neq \beta$, we may suppose that $\alpha > \beta$, so $g(\alpha) \leq g(\beta)$ (since $g$ is non-increasing), so $\alpha \leq \beta$ (since $\alpha$ and $\beta$ are fixpoints of $g$), which is in contradiction with the assumption $\alpha > \beta$.

- if $\mathcal{R}^-(A) \neq \emptyset$ with $\mathcal{R}^-(A) = \{A_1, \ldots, A_n\}$, $Cat(A) = \frac{1}{1+Cat(A_1)+\ldots+Cat(A_n)}$

The translation of $Cat$ into a gradual valuation is: $V = [0,1]$, $W = [0, \infty[$, $V_{\text{Min}} = 0$ and $V_{\text{Max}} = 1$ and $g: W \to V$ is defined by $g(x) = \frac{1}{1+x}$ and $h$ is defined by $h(\{x_1, \ldots, x_n\}) = x_1 + \cdots + x_n$. $\square$

**Proof**
**(of Property 7)** Let $t = (x_1, \ldots, x_n, \ldots)$, $t' = (y_1, \ldots, y_n, \ldots)$, $t'' = (z_1, \ldots, z_n, \ldots)$ be tuples.

**Commutativity of $\star$:** $t \star t' = t' \star t$ There are two cases:
- if $t$ or $t' = 0^\infty$, the property is given by Definition 8.
- if $t$ and $t' \neq 0^\infty$:

$$\begin{aligned} t \star t' &= \mathtt{Sort}(x_1, \ldots, x_n, \ldots, y_1, \ldots, y_n, \ldots) \\ &= \mathtt{Sort}(y_1, \ldots, y_n, \ldots, x_1, \ldots, x_n, \ldots) \\ &= t' \star t \end{aligned}$$

**Associativity of $\star$:** $(t \star t') \star t'' = t \star (t' \star t'')$ There are two cases:
- if $t$ or $t'$ or $t'' = 0^\infty$, we can simplify the expression. For example, if $t = 0^\infty$:

$$\begin{aligned} (t \star t') \star t'' &= t' \star t'' \\ &= t \star (t' \star t'') \end{aligned}$$

- if $t$, $t'$ and $t'' \neq 0^\infty$:

$$\begin{aligned} (t \star t') \star t'' &= \mathtt{Sort}(x_1, \ldots, x_n, \ldots, y_1, \ldots, y_n, \ldots, z_1, \ldots, z_n, \ldots) \\ &= t \star (t' \star t'') \end{aligned}$$

**Property of $\oplus$:** $(t \oplus k) \oplus k' = t \oplus (k + k')$ We have:

$$\begin{aligned} (t \oplus k) \oplus k' &= (x_1 + k, \ldots, x_n + k, \ldots) \oplus k' \\ &= (x_1 + k + k', \ldots, x_n + k + k', \ldots) \\ &= t \oplus (k + k') \end{aligned}$$

**Distributivity:** $(t \star t') \oplus k = (t \oplus k) \star (t' \oplus k)$ We have:

$$\begin{aligned} (t \star t') \oplus k &= \mathtt{Sort}(x_1, \ldots, x_n, \ldots, x'_1, \ldots, x'_n, \ldots) \oplus k \\ &= \mathtt{Sort}(x_1 + k, \ldots, x_n + k, \ldots, x'_1 + k, \ldots, x'_n + k, \ldots) \\ &= (t \oplus k) \star (t' \oplus k) \end{aligned}$$

$\square$

**Proof**
**(of Property 9)** First, we show that the relation $\succeq$ defined by Algorithm 1 is a partial ordering:

Let $u$, $v$, $w$ be three tupled values, the relation $\succeq$ defined by Algorithm 1 is:

- reflexive: $u \succeq u$ because $u = u$, so $u \succeq u$ AND $u \succeq u$ (case 1 of Algorithm 1);
- transitive: suppose that $u \succeq v$ and $v \succeq w$ and consider all the possible cases:
    - if $u = v$:
        - if $v = w$: then $u = w$ so $u \succeq w$,
        - if $|v_i| \leq |w_i|$ AND $|v_p| > |w_p|$: then $|v_i| = |u_i| \leq |w_i|$ AND $|v_p| = |u_p| > |w_p|$, so $u \succeq w$,
        - if $|v_i| < |w_i|$ AND $|v_p| \geq |w_p|$: then $|v_i| = |u_i| < |w_i|$ AND $|v_p| = |u_p| \geq |w_p|$, so $u \succeq w$,
        - if $|v_i| = |w_i|$ AND $|v_p| = |w_p|$ AND $v_p \leq_{lex\infty} w_p$ AND $v_i \geq_{lex\infty} w_i$: then $|v_i| = |u_i| = |w_i|$ AND $|v_p| = |u_p| = |w_p|$ AND $v_p = u_p \leq_{lex\infty} w_p$ AND $v_i = u_i \geq_{lex\infty} w_i$, so $u \succeq w$;
    - if $|u_i| \leq |v_i|$ AND $|u_p| > |v_p|$:
        - if $v = w$: then $|u_i| \leq |v_i| = |w_i|$ AND $|u_p| > |v_p| = |w_p|$ so $u \succeq w$,
        - if $|v_i| \leq |w_i|$ AND $|v_p| > |w_p|$: then $|u_i| \leq |v_i| \leq |w_i|$ AND $|u_p| > |v_p| > |w_p|$, so $u \succeq w$,
        - if $|v_i| < |w_i|$ AND $|v_p| \geq |w_p|$: then $|u_i| \leq |v_i| < |w_i|$ AND $|u_p| > |v_p| \geq |w_p|$, so $u \succeq w$,
        - if $|v_i| = |w_i|$ AND $|v_p| = |w_p|$: then $|u_i| \leq |v_i| = |w_i|$ AND $|u_p| > |v_p| = |w_p|$, so $u \succeq w$;
    - if $|u_i| < |v_i|$ AND $|u_p| \geq |v_p|$:
        - if $v = w$: then $|u_i| < |v_i| = |w_i|$ AND $|u_p| \geq |v_p| = |w_p|$ so $u \succeq w$,
        - if $|v_i| \leq |w_i|$ AND $|v_p| > |w_p|$: then $|u_i| < |v_i| \leq |w_i|$ AND $|u_p| \geq |v_p| > |w_p|$, so $u \succeq w$,
        - if $|v_i| < |w_i|$ AND $|v_p| \geq |w_p|$: then $|u_i| < |v_i| < |w_i|$ AND $|u_p| \geq |v_p| \geq |w_p|$, so $u \succeq w$,
        - if $|v_i| = |w_i|$ AND $|v_p| = |w_p|$: then $|u_i| < |v_i| = |w_i|$ AND $|u_p| \geq |v_p| = |w_p|$, so $u \succeq w$;
    - if $|u_i| = |v_i|$ AND $|u_p| = |v_p|$ AND $u_p \leq_{lex\infty} v_p$ AND $u_i \geq_{lex\infty} v_i$:
        - if $v = w$: then $|u_i| = |v_i| = |w_i|$ AND $|u_p| = |v_p| = |w_p|$ AND $u_p \leq_{lex\infty} v_p = w_p$ AND $u_i \geq_{lex\infty} v_i = w_i$ so $u \succeq w$,
        - if $|v_i| \leq |w_i|$ AND $|v_p| > |w_p|$: then $|u_i| = |v_i| \leq |w_i|$ AND $|u_p| = |v_p| > |w_p|$, so $u \succeq w$,

- if $|v_i| < |w_i|$ AND $|v_p| \geq |w_p|$: then $|u_i| = |v_i| < |w_i|$ AND $|u_p| = |v_p| \geq |w_p|$, so $u \succeq w$,
- if $|v_i| = |w_i|$ AND $|v_p| = |w_p|$ AND $v_p \leq_{lex\infty} w_p$ AND $v_i \geq_{lex\infty} w_i$: then $|u_i| = |v_i| = |w_i|$ AND $|u_p| = |v_p| = |w_p|$ AND $u_p \leq_{lex\infty} v_p \leq_{lex\infty} w_p$ AND $u_i \geq_{lex\infty} v_i \geq_{lex\infty} w_i$, so $u \succeq w$.

In all cases, $u \succeq w$.

Now, consider the maximal and minimal values:

- The tupled value $[0^\infty, ()]$ is the unique maximal element for the preordering $\succeq$: let $v$ be a tupled value such that $v \neq [0^\infty, ()]$, then $|v_p| \leq \infty$ and $|v_i| \geq 0$. Compare $[0^\infty, ()]$ and $v$ with Algorithm 1: $[0^\infty, ()] \neq v$ so the case number 1 is not used; then, $|()| = 0 \leq |v_i|$ AND $|0^\infty| = \infty \geq |v_p|$ so there are two cases:
  - if $|v_p| = \infty$ and $|v_i| = 0$, the case 3 of Algorithm 1 is applied and $[0^\infty, ()] \succ v$,
  - else $|v_p| \leq \infty$ and $|v_i| \geq 0$, the case 5 of Algorithm 1 is applied and $[0^\infty, ()] \succ v$.
- The tupled value $[(), 1^\infty]$ is the unique minimal element for the preordering $\succeq$: let $v$ be a tupled value such that $v \neq [(), 1^\infty]$, then $|v_i| \leq \infty$ and $|v_p| \geq 0$. Compare $[(), 1^\infty]$ and $v$ with Algorithm 1: $[(), 1^\infty] \neq v$ so the case number 1 is not used; then, $|()| = 0 \leq |v_p|$ AND $|1^\infty| = \infty \geq |v_i|$ so there are two cases:
  - if $|v_i| = \infty$ and $|v_p| = 0$, the case 2 of Algorithm 1 is applied and $[(), 1^\infty] \prec v$,
  - else $|v_i| \leq \infty$ and $|v_p| \geq 0$, the case 6 of Algorithm 1 is applied and $[(), 1^\infty] \prec v$.

□

**Proof**
**(of Property 10)** The principle **P1'** is satisfied by Definition 10 and by the fact that $[0^\infty, ()]$ is the unique maximal element of $v(\mathcal{A})$ (see Property 9).

The principle **P2'** is satisfied because of Definition 10.

The principles **P3'** and **P4'** are satisfied: all the possible cases of improvement/degradation of the defence/attack for a given argument (see Definition 16) are applied case by case[31]. Each case leads to a new argument. Using Algorithm 1, the comparison between the argument before and after the application of the case shows that the principle **P3'** (or **P4'**, depending on the applied case)

---

31. We work case by case in order to avoid the complex cases in which we have several simultaneous simple modifications. For example, the modification of the length of a branch which changes the status of the branch (an even integer replaced by an odd integer) is a complex case corresponding to two simple cases: the removal of a branch with a given status, then the addition of a new branch with a different status.

is satisfied. □

**Proof**
**(of Property 11)** From Definition 10. □

**Proof**
**(of Property 12)** First, we consider the case of the preferred extensions: Let $E$ be a preferred extension $\subseteq \mathcal{A}$, we assume that $E$ does not contain all the unattacked arguments of $\mathcal{A}$. So, let $A \in \mathcal{A}$ be an unattacked argument such that $A \notin E$.
Consider $E \cup \{A\}$:

- If $E \cup \{A\}$ is conflict-free then, with $A$ an unattacked argument and $E$ a preferred extension, $E \cup \{A\}$ collectively defends itself, so $E \cup \{A\}$ is admissible and $E \subseteq E \cup \{A\}$. This contradicts the fact that $E$ is a preferred extension.
- If $E \cup \{A\}$ contains at least one conflict, then:
    - $\exists B \in E$ such that $B\mathcal{R}A$. This is impossible since $A$ is unattacked.
    - or $\exists B \in E$ such that $A\mathcal{R}B$. But, since $A$ is unattacked, $\nexists C \in E$ such that $C\mathcal{R}A$. So, $E$ does not collectively defend $B$, which is in contradiction with the fact that $E$ is a preferred extension.

So, the assumption "$E$ does not contain all the unattacked arguments of $\mathcal{A}$" cannot hold.

Now, we consider stable extensions: Let $E$ be a stable extension $\subseteq \mathcal{A}$, we assume that $E$ does not contain all the unattacked arguments of $\mathcal{A}$. So, let $A \in \mathcal{A}$ be an unattacked argument such that $A \notin E$.
Since $A \notin E$ there exists in $E$ another argument $B$ which attacks $A$; This is impossible since $A$ is unattacked.
So, the assumption "$E$ does not contain all the unattacked arguments of $\mathcal{A}$" cannot hold. □

**Proof**
**(of Property 13)** An argument and one of its direct attackers cannot belong to the same extension in the sense of Dung (1995) because the extension must be conflict-free. So, since $A$ is uni-accepted, it means that $A$ belongs to all the extensions, and none of the direct attackers of $A$ belongs to these extensions.

For the converse, we use the following counterexample in the case of the preferred semantics:

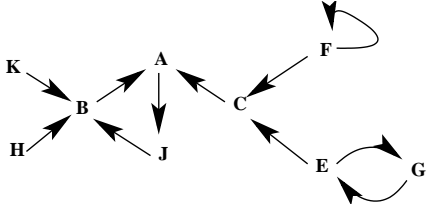

There are two preferred extensions $\{K, H, G\}$ and $\{A, E, K, H\}$. The argument $A$ is cleanly-accepted ($B$ and $C$ do not belong to any preferred extension, and $A$ belongs to at least one of the two extensions). But, $A$ is not uni-accepted because it does not belong to all preferred extensions.

□

**Proof**
**(of Property 14)** First, $A$ uni-accepted $\Rightarrow$ $A$ cleanly-accepted is the result of Property 13.

Conversely, let $A$ be a cleanly-accepted argument, there exists at least one stable extension $E$ such that $A \in E$ and $\forall B, B\mathcal{R}A, B \notin E'$, $\forall E'$ stable extension. Using a *reductio ad absurdum*, we assume that there exists a stable extension $E''$ such that $A \notin E''$; but, if $A \notin E''$, it means that $\exists B \in E''$ such that $B\mathcal{R}A$, so, the direct attacker $B$ of $A$ belongs to a stable extension; so, there is a contradiction with the assumption ($A$ is cleanly-accepted); so, $E''$ does not exist and $A$ is uni-accepted. □

**Proof**
**(of Theorem 1)**

1. We consider the arguments $B \in \mathcal{A}$ such that $B \neq A$. Let $X_i$ be a leaf, the path $\mathcal{C} \in \mathcal{C}(X_i, A)$ is $X_i^1 - \ldots - X_i^{l_i} - A$ with $X_i^1 = X_i$ and $l_i$ denoting the length of the path (if $l_i$ is even, this path is a defence branch for $A$, else it is an attack branch).
   The constraints from $X_i^1$ to $X_i^{l_i}$ are the following:

   $$X_i^1 \succ X_i^3 \succ \ldots \succ X_i^{l_i} \succ X_i^{l_i-1} \succ \ldots \succ X_i^4 \succ X_i^2 \text{ if } l_i \text{ odd } \geq 1$$

   or

   $$X_i^1 \succ X_i^3 \succ \ldots \succ X_i^{l_i-1} \succ X_i^{l_i} \succ \ldots \succ X_i^4 \succ X_i^2 \text{ if } l_i \text{ even } \geq 2$$

   So, for the path $X_i^1 - \ldots - X_i^{l_i}$, the set of the well-defended arguments is $\{X_i^1, X_i^3, \ldots, X_i^{l_i}\}$ if $l_i$ is odd, $\{X_i^1, X_i^3, \ldots, X_i^{l_i-1}\}$ otherwise (this is the set of all the arguments having a value strictly better than those of their direct attackers). This set will denoted by $\text{ACCEP}_i$.

   By definition, this set is conflict-free, it defends all its elements (because it contains only the leaf of the path and all the arguments which are defended by this leaf) and it attacks all the other arguments of the path. If we try to include another argument of the path $X \in \{X_i^1, \ldots, X_i^{l_i}\} \setminus \text{ACCEP}_i$, we obtain a conflict (because **all** the other arguments of the path are attacked by the elements of $\text{ACCEP}_i$). So, for $\{X_i^1, \ldots, X_i^{l_i}\}$, $\text{ACCEP}_i$ is the only preferred and stable extension.

Consider $\mathcal{A}' = \mathcal{A} \setminus \{A\}$, with $\mathcal{R}'$ being the restriction of $\mathcal{R}$ to $\mathcal{A}'$[32] and $\text{UNION\_ACCEP} = \cup_i \text{ACCEP}_i$, then $\text{UNION\_ACCEP}$ is the only preferred and stable extension of $<\mathcal{A}', \mathcal{R}'>$.

So, $\forall B \in \mathcal{A}, B \neq A$, $B$ is accepted iff $B$ well-defended.

2. Now, consider $A$. If $A$ is accepted then $\text{UNION\_ACCEP} \cup \{A\}$ is the only preferred and stable extension of $<\mathcal{A}, \mathcal{R}>$. So, $\forall i$, $X_i^{l_i}$ does not belong to the extension. Then, $\forall i$, $X_i^{l_i-1} \succ X_i^{l_i}$. Therefore, each branch leading to $A$ is a defence branch for $A$. So, $\forall i$, $v(X_i^{l_i}) = [()(l_i - 1)]$. So, $v(A) = [(l_1, l_2, \ldots, l_n)()]$. Then, $\forall i$, $v(A) \succ v(X_i^{l_i})$. Therefore, $A$ is well-defended.

Using the following example, we show that the converse is false:

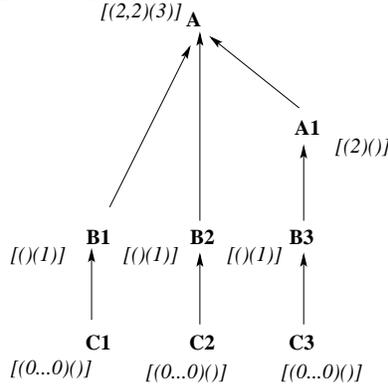

$A$ is well-defended ($A \succ B_1$, $A \succ B_2$ and $A$ is incomparable with $A_1$) but not accepted.

3. Now, if $A$ is well-defended and all the branches leading to $A$ are defence branches for $A$, then $\text{UNION\_ACCEP} \cup \{A\}$ is conflict-free and $A$ is defended against each of its direct attackers (because $X_i^{l_i-1} \in \text{UNION\_ACCEP}$ for each branch $i$). So, $\text{UNION\_ACCEP} \cup \{A\}$ is the preferred and stable extension of $<\mathcal{A}, \mathcal{R}>$ and $A$ is accepted.

$\square$

**Proof**
**(of Lemma 1)** Let $<\mathcal{A}, \mathcal{R}>$ be an argumentation system with a finite relation $\mathcal{R}$ without cycles (so, there is only one non empty preferred and stable extension denoted by $E$). We know that:

- if $A$ is exi-accepted and if $A$ has a direct attacker denoted by $B$ then $B$ is not-accepted,

---

32. $\mathcal{R}'$ is the restriction of $\mathcal{R}$ to $\mathcal{A}'$ if and only if $\mathcal{R}' = \{(a, b) | a\mathcal{R}b, a \in \mathcal{A}', b \in \mathcal{A}'\}$.

- if $B$ is not-accepted then there exists at least one argument $C$ such that $C\mathcal{R}B$ and $C$ is exi-accepted (because $B$ does not belong to $E$ and $E$ is stable, so $C$ must $\in E$). So, *a fortiori*, if $B$ is not-accepted and has only one direct attacker $C$, then $C$ will be exi-accepted.

The proof is done by induction on the depth of a proof tree for $A$ or $C$.

- Basic case for $(i)$: $A$ is exi-accepted with only one direct attacker $B$ ($B\mathcal{R}A$) and $C_1 \ldots C_n$ are the direct attackers of $B$; so, we have a proof tree whose depth is 2 for $A$ and one of the unattacked $C_i$, for example $C_1$; so:

$$\begin{aligned} v(B) &= g(h(v(C_1), \ldots, v(C_n))) \\ &\leq g(v(C_1)) \quad \text{because } h(v(C_1), \ldots, v(C_n)) \geq h(v(C_1)) = v(C_1) \\ &\qquad \text{and } g \text{ is non-increasing} \\ &\leq g(V_{\text{Max}}) \quad \text{because } v(C_1) = V_{\text{Max}} \end{aligned}$$

so:

$$\begin{aligned} v(A) &= g(v(B)) \\ &\geq g^2(V_{\text{Max}}) \end{aligned}$$

But, Property 1 says that $g^2(V_{\text{Max}}) \geq g(V_{\text{Max}})$, so $v(A) \geq v(B)$.

- Basic case for $(ii)$: $C\mathcal{R}B$ with $C$ the only direct attacker of $B$; so, we have a proof tree whose depth is 0 for $C$, *i.e.* $C$ is unattacked; so, $v(C) = V_{\text{Max}}$ and $v(B) = g(V_{\text{Max}}) \leq v(C)$ (following Definition 6).

- General case for $(i)$: $A$ is exi-accepted with only one direct attacker $B$ ($B\mathcal{R}A$) and $C_1 \ldots C_n$ are the direct attackers of $B$, with one of the $C_i$ exi-accepted, for example $C_1$; we consider the subgraph leading to $C_1$ to which we add $C_1\mathcal{R}B\mathcal{R}A$, and we assume:
  $g(v(C_1)) \leq v(C_1)$ (induction assumption issued from $(ii)$)
So:

$$\begin{aligned} v(B) &= g(h(v(C_1), \ldots, v(C_n))) \\ &\leq g(v(C_1)) \quad \text{for the same reasons as in the basic case} \\ &\leq v(C_1) \quad \text{by induction assumption} \\ &\leq h(v(C_1), \ldots, v(C_n)) \quad \text{property of } h \end{aligned}$$

and with the non-increasing of $g$:

$$\begin{aligned} v(A) &= g(v(B)) \\ &\geq g(h(v(C_1), \ldots, v(C_n))) = v(B) \end{aligned}$$

- General case for (ii): $B$ is not-accepted, so $C$ is exi-accepted; we assume that $C$ has several direct attackers $D_1 \ldots D_p$ which are all not-accepted (because $C$ is exi-accepted); we consider each subgraph leading to $D_i$ to which we add $D_i \mathcal{R} C \mathcal{R} B$ and we assume:
  $\forall i = 1 \ldots p, g(v(D_i)) \geq v(D_i)$ (induction assumption issued from (i))
  so:

$$\begin{aligned} v(C) &= g(h(v(D_1), \ldots, v(D_p))) \\ &\geq h(v(D_1), \ldots, v(D_p)) \quad \text{application of the condition } (*) \\ & \quad \text{since the induction assumption} \\ & \quad \text{corresponds to the premise of } (*) \end{aligned}$$

so:

$$\begin{aligned} v(B) &= g(v(C)) \\ &\leq g(h(v(D_1), \ldots, v(D_p))) = v(C) \end{aligned}$$

$\square$

**Proof**
**(of Theorem 2)** Assume that $(*)$ is true and consider $A \in \mathcal{A}$ which is exi-accepted. Let $B_i$, $i = 1 \ldots n$, be the direct attackers of $A$. Then, for all $i = 1 \ldots n$, in the subgraph leading to $B_i$ and completed with $B_i \mathcal{R} A$, we apply the lemma and we obtain: $g(v(B_i)) \geq v(B_i)$, $\forall i = 1 \ldots n$. Thus, we have:

$$\begin{aligned} v(A) &= g(h(v(B_1), \ldots, v(B_n))) \\ &\geq h(v(B_1), \ldots, v(B_n)) \quad \text{by applying } (*) \\ &\geq v(B_i), \forall i = 1 \ldots n \quad \text{property of } h \end{aligned}$$

So, $A$ is well-defended.

For the converse, let $A \in \mathcal{A}$ be well-defended. Let $B_1, \ldots, B_n$ be the direct attackers of $A$ and assume that $A$ is not exi-accepted. Then, there exists at least one direct attacker $B_i$ of $A$ such that $B_i$ is exi-accepted (because there is only one preferred and stable extension). We can apply (ii) of the lemma on the subgraph leading to $B_i$ completed with $B_i \mathcal{R} A$ and we obtain $g(v(B_i)) \leq v(B_i)$. So, there exists $B_i$ a direct attacker of $A$ such that:

$$\begin{aligned} v(A) &= g(h(v(B_1), \ldots, v(B_n))) \\ &\leq g(v(B_i)) \quad \text{property of } h \text{ and non-increasing of } g \\ &\leq v(B_i) \quad \text{using the lemma} \end{aligned}$$

This is in contradiction with $A$ well-defended. So, $A$ is exi-accepted. □

## Appendix B. Computation of tupled values

We propose an algorithm for computing the tupled values for an arbitrary graph (cyclic or acyclic, the cycles may be isolated or not). This algorithm uses a principle of propagation of values: an argument is evaluated when the values of its direct attackers are known.

We must consider the cycles as meta-arguments which are evaluated when all the "direct attackers of the cycle" (*i.e.* the direct attackers of one of the elements of the cycle which do not belong to the cycle) are evaluated.

The beginning of the process is as follows: we consider that all the arguments have the initial value $[0^\infty, ()]$, and only the leaves of the graph are "marked" as having their final values. Thus, we have the following partition of the graph $\mathcal{G}$:

- $\mathcal{G}_v$: the part of the graph already evaluated (at the beginning, this part contains only the leaves of the graph),
- $\mathcal{G}_{\neg v}$: the part of the graph which is not evaluated (at the beginning, this part contains all the arguments of the graph $\mathcal{G}$ except the leaves).

The algorithm also relies on a special data structure denoted by $\mathcal{L}$ giving the list of the cycles in the graph and their main characteristics:

- list of the arguments which belong to this cycle,
- list of the arguments which belong to this cycle and which have direct attackers outside the cycle (these arguments are called *inputs of the cycle*; those which will be used in order to propagate the values across the cycle in the case of a non isolated cycle); this list will be empty in the case of an isolated cycle.

**Remark:** For the sake of efficiency, the interconnected cycles (see Definition 1) will be considered as a "whole" by the algorithm and will be used like a "meta-cycle". For example, the two cycles $A - B - A$ and $B - C - B$ which do not have any direct attacker outside of the cycles, will be described in the data structure $\mathcal{L}$ as only one "meta-cycle" with the following lists:

- $A$, $B$, $C$,
- nothing (because it is an isolated "meta-cycle").

In order to avoid some ambiguity, these "meta-cycles" are defined as *mcycles*:

**Definition 21 (mcycle)** *Let $\mathcal{G}$ be an attack graph. Let $\mathcal{CC}$ be the set of all the cycles of $\mathcal{G}$. Let $\mathcal{CC}' \subseteq \mathcal{CC}$ and $\mathcal{CC}' = \{\mathcal{C}_1, \ldots, \mathcal{C}_n\}$ be a set of cycles.*
*Let $\mathcal{A}_{\mathcal{CC}'}$ be the set: $\{A_j$ such that $\exists \mathcal{C}_i \in \mathcal{CC}'$ and $A_j \in \mathcal{C}_i\}$.*
*If $\mathcal{CC}'$ satisfies the following properties:*

- $\forall A_j, A_k \in \mathcal{A}_{\mathcal{CC}'}$, $\exists$ *a path from $A_j$ to $A_k$ such that each element (arguments or edges between arguments) of the path belongs to cycles of $\mathcal{CC}'$,*
- *and $\forall \mathcal{C}_k \in \mathcal{CC} \setminus \mathcal{CC}'$, $\nexists \mathcal{C}_i \in \mathcal{CC}'$ such that $\mathcal{C}_k$ is interconnected with $\mathcal{C}_i$.*

*Then the union of the $C_i$ belonging to $CC'$ is a* mcycle.

Thus, we make a partition of $CC$ using the notion of interconnection between cycles, each element of the partition being a different *mcycle*. See on the following example:

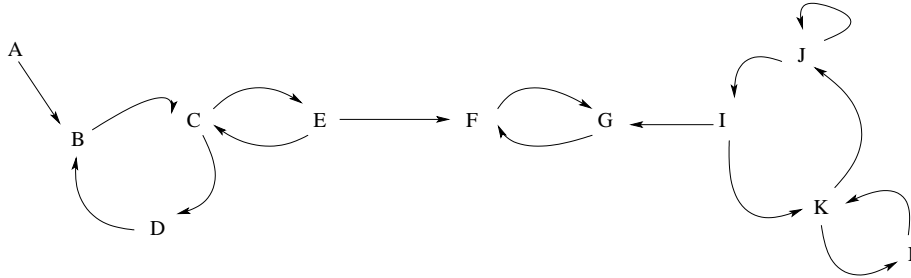

In this graph, there are 6 cycles:

- $\{J\}$,
- $\{I, J, K\}$,
- $\{K, L\}$,
- $\{B, C, D\}$,
- $\{C, E\}$,
- $\{F, G\}$.

and 3 mcycles:

- $\{I, J, K, L\}$,
- $\{B, C, D, E\}$,
- $\{F, G\}$.

Algorithm 2 is the main algorithm used for computing the tupled values.
The function ADD-NODE (respectively REMOVE-NODE) whose parameters are a subgraph $\mathcal{G}_x$ of the attack graph and a node $s$, adds (resp. removes) $s$ in (resp. of) $\mathcal{G}_x$. The other functions are described in (Cayrol & Lagasquie-Schiex, 2003a).
Algorithm 2 has been applied on an example after the step of rewriting (see Figure 1). Note that in order to make the understanding of the results easier, we do not have created new arguments (as in Definitions 11 and 12), but of course, it would be necessary for a rigorous formalization.

**Algorithm 2:** Algorithm for computing tupled values

| | |
|---|---|
| % **Description of parameters:** | % |
| %     $\mathcal{G}$: attack graph (partitioned in $\mathcal{G}_v$ and $\mathcal{G}_{\neg v}$) | % |
| %     $\mathcal{L}$: data structure describing the mcycles | % |
| %     $n$: number of propagation steps for the mcycles | % |
| % **Used variables:** | % |
| %     $A$: the current argument (to be evaluated) | % |
| %     $\mathcal{C}$: the current mcycle (to be evaluated) (containing $A$) | % |
| %     $LAD$: list of the direct attackers of $\mathcal{C}$ | % |
| %     $B_i$: the current direct attackers of $A$, or of $\mathcal{C}$ | % |

**begin**
1  **while** *there is at least one argument in $\mathcal{G}_{\neg v}$* **do**
2    $A = $ Choose-Argument$(\mathcal{G}_{\neg v})$
3    **if** *$A$ does not belong to a mcycle $\mathcal{C}$ described in $\mathcal{L}$* **then**
4      **if** $\forall B_i \in \mathcal{R}^-(A)$, $B_i$ *is already evaluated* **then**
5        $\mathcal{G}_v = $ Add-Node$(\mathcal{G}_v,$Evaluate-Node$(A, \mathcal{R}^-(A), 1))$   % The value of $A$ %
                                                                                        % is the value of its %
                                                                                        % direct attackers %
                                                                                        % in which we add 1 %
                                                                                        % see Definition 10 %
6        $\mathcal{G}_{\neg v} = $ Remove-Node$(\mathcal{G}_{\neg v}, A)$

7    **else**
8      **if** *$\mathcal{C}$ is isolated* **then**
9        $\mathcal{G}_v = $ Add-Mcycle$(\mathcal{G}_v,$Evaluate-Mcycle-Isolated$(\mathcal{G}, \mathcal{C}, n))$
10       $\mathcal{G}_{\neg v} = $ Remove-Mcycle$(\mathcal{G}_{\neg v}, \mathcal{C})$
11     **else**
12       $LAD = $ Find-Direct-Attackers-Mcycle$(\mathcal{C}, \mathcal{G})$
13       **if** $\forall B_i \in LAD$, $B_i$ *is already evaluated* **then**
14         $\mathcal{G}_v = $ Add-Mcycle$(\mathcal{G}_v,$
                   Evaluate-Mcycle-Not-Isolated$(\mathcal{G}, \mathcal{C}, LAD, n))$
15         $\mathcal{G}_{\neg v} = $ Remove-Mcycle$(\mathcal{G}_{\neg v}, \mathcal{C})$

16  **return** $\mathcal{G}$
**end**

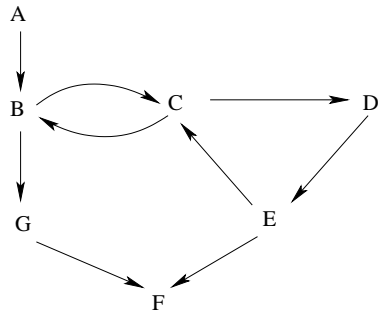

The previous argumentation graph can be rewritten as follows:

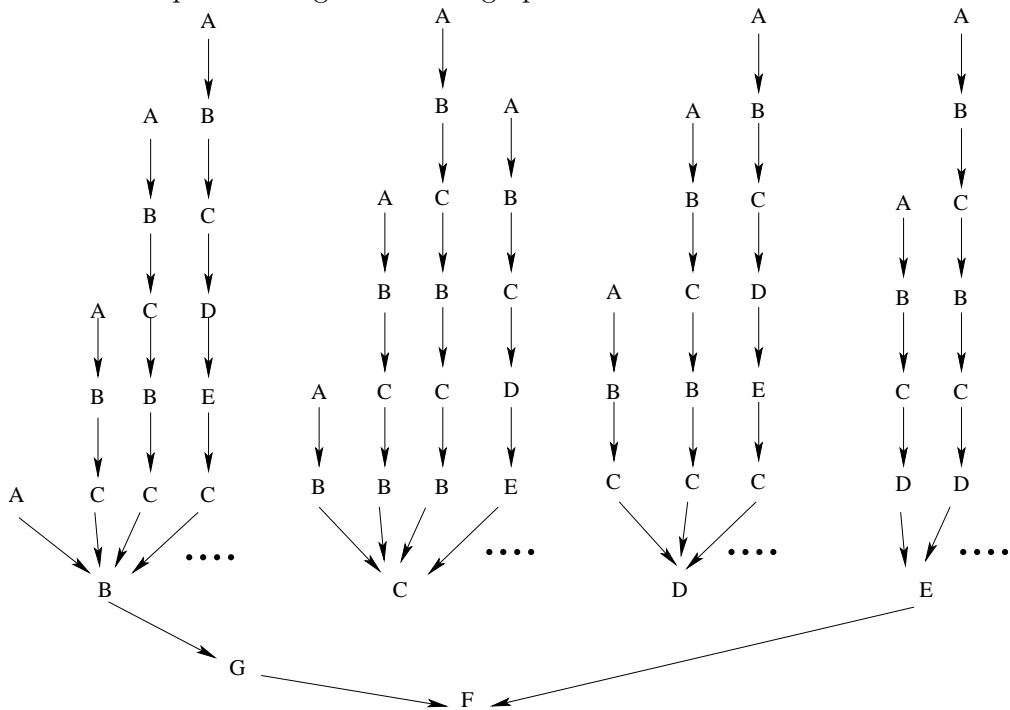

The results of the valuation obtained after one propagation step are:

- $v(A) = [(0, \ldots, 0)()]$,
- $v(B) = [(6, 8, 8, \ldots)(1, 3, 5, \ldots)]$,
- $v(C) = [(2, 4, 6, \ldots)(5, \ldots)]$,
- $v(D) = [(6, \ldots)(3, 5, \ldots)]$,
- $v(E) = [(4, 6, \ldots)()]$,
- $v(F) = [(8, \ldots)(3, 5, 5, 7, 7, \ldots)]$,
- $v(G) = [(2, 4, 6, \ldots)(7, \ldots)]$,

Figure 1: Example of rewriting